\pdfoutput=1
\documentclass[11pt, letterpaper, logo, onecolumn, copyright]{template_from_google}
\usepackage{float}
\usepackage{wrapfig}
\usepackage[T1]{fontenc}
\usepackage{CJKutf8} 
\usepackage{amsmath}
\usepackage{amsfonts}
\usepackage{amssymb}
\usepackage{amsthm}
\usepackage{bm} 
\usepackage{nicefrac} 

\usepackage{booktabs} 
\usepackage{enumitem} 
\usepackage{fancyhdr}
\usepackage{multirow} 
\usepackage{tabularx}
\usepackage{multicol}
\usepackage{adjustbox}
\usepackage{array}
\usepackage{longtable}
\usepackage{lipsum} 
\usepackage{wrapfig} 
\usepackage{tocloft} 
\usepackage{fancybox}
\usepackage{authblk} 

\usepackage{graphicx}
\usepackage{caption}
\usepackage{subcaption}
\usepackage{capt-of} 
\usepackage[inkscapeformat=png]{svg}

\usepackage{listings, listings-rust}
\usepackage{listingsutf8}
\usepackage[ruled,vlined]{algorithm2e}

\usepackage[dvipsnames]{xcolor}
\usepackage[most, breakable, skins]{tcolorbox}
\tcbuselibrary{listings,skins,breakable}

\usepackage{fontawesome5}
\usepackage{xspace} 

\usepackage[authoryear, sort&compress, round]{natbib}

\usepackage[dvipsnames]{xcolor} 

\theoremstyle{plain}

\theoremstyle{definition}

\theoremstyle{remark}

\definecolor{darkblue}{rgb}{0.0, 0.0, 0.6}
\definecolor{darkred}{rgb}{0.7, 0.0, 0.0}
\hypersetup{
  pdffitwindow=true,
  pdfstartview={FitH},
  pdfnewwindow=true,
  colorlinks,
  linktocpage=true,
  linkcolor=darkred,
  urlcolor=darkblue,
  citecolor=darkblue
}
\usepackage{hyperref}

\makeatletter

\renewcommand\paragraph{\@startsection{paragraph}{4}{\z@}%
            {-2.5ex\@plus -1ex \@minus -.25ex}%
            {1.25ex \@plus .25ex}%
            {\itshape\normalsize\bfseries}}
\makeatother

\let\cite\citep

\title{BLADE: Bayesian Langevin Active Discovery with Replica Exchange for Identification of Complex Systems}
\newcommand{\shorttitle}{BLADE}
\reportnumber{} 

\renewcommand{\today}{}

\author[1]{Cindy Xiangrui Kong}
\author[1]{Haoyang Zheng}
\author[1,2]{Guang Lin}

\affil[1]{Department of Mathematics, Purdue University, West Lafayette, IN 47907, USA}

\affil[2]{School of Mechanical Engineering, Purdue University, West Lafayette, IN 47907, USA}

\begin{abstract}
Traditional methods for system discovery frequently struggle with efficient data usage and uncertainty quantification. Identifying the governing equations of complex dynamical systems from data presents a significant challenge in scientific discovery, especially when high-quality measurements are scarce and expensive to obtain. To overcome these limitations, we propose \textbf{B}ayesian \textbf{L}angevin \textbf{A}ctive \textbf{D}iscovery with Replica \textbf{E}xchange for Identification of Complex Systems (BLADE), a novel Bayesian framework that combines replica-exchange stochastic gradient Langevin Monte Carlo with active learning. By balancing gradient-driven exploration and exploitation in coefficient space, BLADE provides probabilistic parameter estimation and principled uncertainty quantification. Faced with data scarcity, the probabilistic foundation of BLADE further facilitates the integration of active learning through a hybrid acquisition strategy that combines predictive uncertainty with space-filling design, enabling efficient selection of informative samples. Across benchmark systems, BLADE reduces measurement requirements by roughly 60\% for Lotka–Volterra and 40\% for Burgers’ equation relative to random sampling, demonstrating substantial data-efficiency gains. These results highlight BLADE as a general uncertainty-aware framework for discovering interpretable dynamical systems, particularly valuable when high-fidelity data acquisition is prohibitively expensive.
\end{abstract}

\begin{document}

\maketitle

\section{Introduction}
Identification of governing equations of complex dynamical systems from data is a longstanding goal in science and engineering. These governing equations, expressed as accurate dynamical models, underpin predictive capability and scientific understanding in domains such as turbulence modeling \citep{Pope2000}, climate dynamics \citep{drake2014}, systems biology \citep{alon2006}, and neuroscience \citep{izhikevich2007}. While many systems are known to follow differential equations, deriving their precise forms from first principles can be intractable due to complexity or incomplete knowledge. 
In such settings, data-driven approaches provide a complementary pathway: they reconstruct the underlying dynamics from measurements, enabling accurate simulation and forecasting.

In recent years, there has been a surge of work on learning nonlinear governing equations from data, which mainly follows two complementary tracks.
First, deep-learning-based formulations approximate the dynamics as universal function approximators, where deep neural network frameworks combined with numerical integration schemes provide black-box approaches to approximate nonlinear differential equations from observed data.
Qin et al. \citep{Qin2019} and Raissi et al. \citep{Raissi2018} use multi‐step and integral formulations to approximate unknown ordinary differential equations, while Goyal and Benner \citep{Goyal2021} embed a fourth‐order Runge–Kutta constraint to accommodate noisy and sparse observations. Gao et al.\citep{Gao2020} employ a nonintrusive reduced‐order approach to tackle large-scale systems described by partial differential equations, and Wu and Xiu \citep{Wu2020}, along with Chen et al. \citep{Chen2022} use evolution operators in modal or nodal spaces to model time‐dependent PDEs. Chen et al. \citep{chen2018neural} introduce continuous-depth models known as neural ordinary differential equations, and deep backward schemes applied by Germain et al. \citep{germain2022approximation} together with high-dimensional approaches from Han et al. \citep{Han2018} address complex PDEs. 
These deep models emphasize predictive accuracy but often sacrifice interpretability and often uncertainty awareness. In addition to neural formulations, probabilistic nonparametric approaches such as Gaussian process vector fields provide calibrated uncertainty estimates while learning unknown ODE dynamics from sparse observations \citep{Heinonen2018}, though interpretability in terms of concise symbolic structure may not be explicit.
A second class of methods, built around sparse regression and symbolic discovery, seeks interpretable structure rather than purely predictive power. There are data-driven approaches that learn interpretable models balancing prediction with scientific insight, among which the Sparse Identification of Nonlinear Dynamics (SINDy) \citep{brunton2016discovering, rudy2019data} framework has become a cornerstone. SINDy assumes that governing dynamics can be expressed as a sparse linear combination of candidate functions, and employs sparse regression to identify the active terms. It has emerged as a transformative paradigm for extracting governing equations from observational data, providing insights into complex phenomena in physics \citep{brunton2016discovering}, biology \citep{zheng2022data}, and engineering \citep{kaheman2020SINDy}. Its success stems from combining interpretability, parsimony, and flexibility in library design.

A critical step in system discovery is the quantification of uncertainty to assess model reliability and compare competing hypotheses. The standard SINDy framework yields a point estimate of the governing equations; in this work, we extend this approach to provide an approximate distribution over the parameters within the selected sparse subspace, thus quantifying the structural and parametric uncertainty. To extend the point estimation of SINDy with a parametric uncertainty guarantee, several extensions have been proposed. Ensemble-based approaches, like Ensemble SINDy \citep{Fasel_2022}, improve robustness by aggregating multiple regressions to provide uncertainty estimates, which can also guide active learning strategies. However, these ensemble-based uncertainties are heuristic and do not correspond to formal posterior distributions. In contrast, Bayesian variants such as UQ-SINDy \citep{hirsh2022sparsifying} offer a more principled probabilistic perspective, using sparsity-promoting priors to infer coefficient posteriors. Nevertheless, these methods are often highly sensitive to prior assumptions and can struggle to capture complex, multi-modal posterior landscapes \citep{fung2024}. Consequently, current frameworks either rely on heuristic ensembles lacking statistical grounding or on Bayesian samplers constrained by prior design and sampling inefficiency. Recent extensions further combine sparse discovery with probabilistic or generative modeling to represent stochasticity in governing equations \citep{jacobs2023hypersindy}, highlighting growing interest in uncertainty-aware and distributional system identification beyond point estimates.

Well-calibrated UQ through the Bayesian framework also suggests a natural mechanism for experimental design: in resource-constrained scenarios, one can actively exploit uncertainty to guide data acquisition. This motivates the use of active learning (AL), where strategies iteratively select informative measurements to minimize data costs.
Traditional AL approaches prioritize either uncertainty reduction (e.g., sampling where predictive variance is high) \citep{MacKay1992, gramacy2009adaptive} or space-filling designs \citep{loeppky2010,yu2010} (e.g., ensuring broad coverage of the state space). Each approach alone has limitations: uncertainty-driven sampling risks over-exploiting local regions, while space-filling designs 
may waste effort in regions already well understood. For dynamical systems, informativeness is often heterogeneous — states near bifurcations or transients may yield far more insight than equilibrium regions \citep{chen2020}. 
These challenges motivate hybrid acquisition strategies that combine uncertainty and diversity, ensuring that new samples both resolve ambiguity and enrich dynamical coverage. On the other hand, practical Bayesian discovery frameworks remain limited: scalable posterior sampling is challenging, and existing AL strategies are rarely integrated with full posterior exploration. As a result, most current methods either provide uncertainty without adaptive acquisition \citep{fung2024}, or apply AL heuristically without calibrated uncertainty \cite{Fasel_2022}. Thus, an integrated Bayesian-active-learning paradigm capable of both calibrated uncertainty quantification and efficient data acquisition remains elusive.

In the face of the above-mentioned challenges, we propose Bayesian Langevin Active Discovery with Replica Exchange for Identification of Complex Systems (BLADE), a unified framework that integrates posterior sampling with hybrid active learning. BLADE leverages replica-exchange stochastic gradient Langevin dynamics (reSGLD) \citep{deng2020non, zheng2024constrained} to sample from the posterior distribution of coefficients.
By combining parallel sampling chains at multiple temperatures, reSGLD enables aggressive exploration of identified coefficient space through high-temperature chains, while low-temperature chains refine high-probability regions. This balance is critical for identifying sparse and interpretable dynamical systems from observational data and providing robust UQ. In contrast to UQ-SINDy, which relies heavily on prior design, BLADE achieves scalable posterior exploration through stochastic gradient sampling, which will be further illustrated in Section 4.1.
Building on this probabilistic foundation, BLADE introduces a hybrid acquisition function that unites predictive uncertainty and space-filling design, systematically reducing data requirements while maintaining accuracy.
Our contributions are threefold:
\begin{description}
  \item[Robust Bayesian Uncertainty Quantification for System Discovery.]
  BLADE leverages reSGLD to efficiently explore high-dimensional coefficient spaces, which overcomes limitations of traditional MCMC methods in exploring multi-modal landscapes and thus enables accurate posterior estimation. 
  \item[Data Efficient Hybrid Active Learning Strategy.]
  A combined predictive-variance + maximin-distance criterion guides measurement selection, cutting data requirements by 40–60\% across benchmark systems.
  \item[Empirical Validation Across Multiple Dynamical Systems.]
BLADE achieves state-of-the-art performance in accuracy, uncertainty calibration, and model sparsity across the Lotka–Volterra and Lorenz systems, as well as Burgers’ and convection–diffusion equations. The robustness of the framework is further demonstrated through efficient AL-driven discovery in limited-data regimes.
\end{description}
This work bridges the gap between probabilistic inference and data efficiency, offering a principled tool for scenarios where high-fidelity data are prohibitively expensive. 

The remainder of this paper is organized as follows: Section 2 formulates the problem of data-driven dynamical system discovery. Section 3 introduces the BLADE framework, which details its integration of reSGLD for uncertainty-aware model identification and an AL strategy that leverages its UQ capabilities to optimize data acquisition in the face of data scarcity. Section 4 validates the framework through experiments on four non-linear systems: the Lotka-Volterra system, the Lorenz system, the Burgers' equation, and the convection–diffusion equation, which demonstrates robustness to noise. In Section 5, we extend the experiments with AL, with benchmarks on the Lotka-Volterra and Burgers’ systems under limited-data regimes. Finally, Section 6 discusses the broader implications, limitations, and future directions.
\section{Problem Statement}
\subsection{Sparse Dynamical System Discovery
}
We study a class of autonomous differential equations given by:
\begin{equation}\label{eq:system}
\frac{\mathrm d}{\mathrm dt} \mathbf{u}(t) = \mathbf{f}(\mathbf{u}(t)),
\end{equation}
where the state vector is defined as $\mathbf{u}(t) = [u_1(t), u_2(t), \cdots, u_d(t)]^\intercal \in \mathbb{R}^d$, and $t$ denotes the corresponding temporal inputs \footnote{For simplicity, we exclude spatial inputs from the formulation \eqref{eq:system}. It should be noted that such extensions could easily extend to partial derivatives with respect to spatial variables. }. 
The dynamics of the system are encoded by the function $\mathbf{f}(\cdot)$, which is unknown and usually of non-linear form.

For the purpose of identifying the system $\mathbf{f}(\cdot)$, data is collected at a sequence of discrete time instances $\{t_n\}_{n=1}^N$. The resulting measurements are aggregated into the matrix $\mathbf{U}=[\mathbf{u}(t_1), \mathbf{u}(t_2), \cdots, \mathbf{u}(t_N)]^\intercal \in \mathbb{R}^{N\times d}$. 
The learning of $\mathbf{f}(\cdot)$ is accomplished by assuming that the underlying dynamical system can be expressed as a linear combination of functions from a candidate library $\bm{\Theta}(\mathbf{U}) = \begin{bmatrix} \theta_1(\mathbf{U}), \cdots, \theta_m(\mathbf{U}) \end{bmatrix} \in \mathbb{R}^{N \times m}$:
\begin{equation}
\dot{\mathbf{U}} = \mathbf{\Theta}(\mathbf{U}) \mathbf{\Xi},
\label{eq: linear}
\end{equation}
where $\dot{\mathbf{U}} \in \mathbb{R}^{N\times d}$ denotes the temporal derivative of the $\mathbf{U}$, $\mathbf{\Xi} \in \mathbb{R}^{m \times d}$ are coefficients that determine which specific functions from the library $\mathbf{\Theta}(\mathbf{U})$ contribute to the system dynamics.
The library of candidate functions is typically constructed from terms known to appear in canonical models of dynamical systems. A common choice is a polynomial basis of the state variables, which, for PDEs, is augmented with partial derivatives. Fourier-based libraries incorporating trigonometric terms, such as $\sin(\mathbf{x})$ and $\cos(\mathbf{x})$, are also frequently employed, especially for systems with periodic behavior.

One key assumption to characterize the system is that the coefficient matrix $\mathbf{\Xi}$ is sparse, i.e., the system can be represented by a small number of candidates in the library $\mathbf{\Theta}$. This is physically justified, as the canonical governing equations describing most natural phenomena are themselves sparse, comprising only a few essential terms. 
The standard SINDy framework \citep{brunton2016discovering} performs sparse system identification using regularized regression, most commonly via Lasso or sequentially thresholded least squares (STLSQ). In this formulation, the discovery problem is decomposed into \(d\) regression tasks, one for each state derivative. This idea was later extended to PDE discovery (PDE-FIND), where sparse coefficients are identified from a candidate library of spatial and nonlinear terms using sequential thresholding with ridge regression \citep{rudy2019data}.
In practice, the sequential thresholding step is used to promote parsimony by iteratively removing coefficients with small magnitude and refitting the remaining active bases. This procedure is effective for recovering sparse governing equations and has well-studied convergence behavior in related settings \citep{zhang2019convergence}. However, it is fundamentally a deterministic sparsification mechanism and, by itself, does not provide posterior uncertainty over the inferred coefficients.
To incorporate uncertainty quantification, we build on this sparse-identification framework and introduce a Bayesian treatment of the coefficient inference problem. In our approach, posterior sampling is used to characterize uncertainty in the coefficients, while sequential thresholding is retained as a practical support-selection step for model parsimony. To enable rigorous uncertainty quantification and robust inference, we reformulate the identification of system \eqref{eq:system} within a Bayesian framework as follows.


\subsection{
Probabilistic Framework for System Discovery
}
A Bayesian framework can be applied to identify and estimate the uncertainty of the systems (\ref{eq:system}), or specifically, we aim to infer the posterior distribution of $\mathbf{\Xi}$.
Let $p(\mathbf{\Xi})$ denote the prior distribution and $p(\mathbf{U}|\mathbf{\Xi})$ the likelihood of the observed data $\mathbf{U}$ given the system parameterized by $\mathbf{\Xi}$. Following the Bayes formula, the posterior distribution of $\mathbf{\Xi}$ given the $N$ collected data snapshots $ \mathbf{U}$ is: 
\begin{equation}
    p(\mathbf{\Xi} | \mathbf{U}) \propto p(\mathbf{\Xi}) \prod_{i=1}^N p(\mathbf{u}(t_i) | \mathbf{\Xi}),
    \label{eq:posterior}
\end{equation} 
where $p(\mathbf{u}(t_i) | \mathbf{\Xi})$ denotes the likelihood for a single observation.
The computation of the posterior distribution $p(\mathbf{\Xi} | \mathbf{U})$ typically presents analytical intractability. In such cases, sampling-based approaches, such as Markov Chain Monte Carlo (MCMC) methods, provide a viable solution. The approximated posterior distribution enables model reconstructions and trajectory forecasts given observational data, allowing quantitative uncertainty assessment of model coefficients and predicted trajectories.

Within the Bayesian framework, it necessitates a prior distribution $p(\mathbf{\Xi})$ that promotes sparsity while robustly handling coefficients of varying magnitudes. A variety of sparsity-promoting priors have been developed, such as the Laplace (Lasso) prior \citep{gelman2013bayesian}, the spike-and-slab prior \citep{ishwaran2005spike}, or the more flexible Regularized Horseshoe prior \citep{Piironen_2017}. These priors apply shrinkage to irrelevant coefficients while preserving the magnitude of important ones, thereby balancing parsimony and robustness. In particular, the Regularized Horseshoe prior is particularly well-suited for identifying parsimonious dynamical systems, as it combines strong shrinkage for irrelevant coefficients with controlled regularization for significant one; moreover, it avoids the over-shrinkage problem by allowing significant coefficients to escape toward a slab component \citep{hirsh2022sparsifying}.

Notably, sparsifying priors alone do not fully guarantee model parsimony. We additionally employ a thresholding step that prunes small-magnitude coefficients after posterior inference. This two-stage approach - probabilistic shrinkage followed by deterministic thresholding - ensures that only the most relevant basis functions are retained, yielding sparse yet reliable governing equations.

\section{Methodology}
In this work, we introduce BLADE, which incorporates Langevin MCMC methods into the discovery of physical laws in a Bayesian way via the sequential thresholding method. BLADE utilizes the stochastic gradient in the situation of sufficient data collection to improve efficiency and can be exploited for AL in the face of data scarcity (i.e., the time derivative calculation or collection is expensive). We start with the data collected $ \mathbf{U} \in \mathbb{R}^{N \times d}$ from the ODE or PDE systems as described before, with the task of identifying the system in terms of (\ref{eq:system}) and determining the posterior distribution of the model coefficient $\mathbf{\Xi} \in \mathbb{R}^{m \times d}$, where $m$ is the number of candidates in each row of the library $\mathbf{\Theta}(\mathbf{U})$. 

\subsection{Uncertainty Quantification with Langevin MCMC}

For ease of notation, we focus on the column vector $\bm{\xi} \in \mathbb{R}^m$ of the model coefficient $\bm{\Xi} \in \mathbb{R}^{m \times d}$, which represents the learned coefficients for one dimension of the system states. This can be easily extended to the sampling of $\bm{\Xi}$ by matching the dimension of the noise term $\bm{\zeta}$ to be introduced. 

Langevin-type algorithms are grounded in Langevin diffusion, a stochastic process governed by the stochastic differential equation (SDE):  
\begin{equation}
d\bm{\xi}_t = -\nabla L(\bm{\xi}_t) \, dt + \sqrt{2\tau} \, d\mathbf{W}_t,
\label{eq:langevin_sde}
\end{equation}
where $L(\cdot)$ represents the energy function, $\{\mathbf{W}_t \mid t \geq 0\}$ denotes the standard Brownian motion on $\mathbb{R}^m$, and $\tau > 0$ is the temperature parameter. In \eqref{eq:langevin_sde}, $t$ denotes the continuous algorithmic time of the diffusion process, rather than the physical time of the dynamical system. Under mild conditions on $L$, it is well-established that the diffusion process described by (\ref{eq:langevin_sde}) admits a unique strong solution $\{\bm{\xi}_t, t \geq 0\}$, which is a Markov process. Moreover, as $t \to \infty$, the distribution of $\bm{\xi}_t$ converges to the invariant distribution $\pi_\tau$, characterized by the density $\pi_\tau(\bm{\xi}) \propto \exp(-L(\bm{\xi}) / \tau)$.  

To numerically approximate Langevin diffusion (\ref{eq:langevin_sde}), the forward Euler discretization is commonly employed. This leads to the iterative update:
\begin{equation}
\tilde{\bm{\xi}}_{k+1} = \tilde{\bm{\xi}}_k - \eta_k \nabla L(\tilde{\bm{\xi}}_k) + \sqrt{2 \eta_k \tau} \, \bm{\zeta}_k,
\label{eq:ULA}
\end{equation}
where 
$\tilde{\bm{\xi}}_k$
indicates a discrete-time approximation of the continuous Langevin diffusion at the $k$-th iteration,
$\bm{\zeta}_k \sim \mathcal{N}(0, \bm{I}_m)$
represents independent Gaussian noise,
and $\eta_k$ is the step size. 

To enable scalable and efficient sampling while maintaining convergence to the target distribution under appropriate conditions, we further consider its stochastic gradient version for efficient large-scale data applications. Specifically, consider the following energy function $\tilde{L}$ (negative logarithmic posterior) approximated with a mini-batch of data $\mathbf{B}$ uniformly subsampled from the given data $\mathbf{U}$ \citep{welling2011bayesian}: 

\begin{equation}
\tilde{L}(\tilde{\bm{\xi}}) = -\log p(\tilde{\bm{\xi}}) - \frac{N}{|\mathbf{B}|} \sum_{\mathbf x_i\in \mathbf{B}} \log P(\mathbf{x}_i \mid \tilde{\bm{\xi}}).
\label{eq:neg-log-post-est}
\end{equation}  
Under the problem setup (\ref{eq: linear}), the gradient $\nabla\tilde{L}(\tilde{\bm{\xi}})$ is calculated as
\begin{equation}
\nabla\tilde{L}(\tilde{\bm{\xi}}) = -\nabla\log p(\tilde{\bm{\xi}}) - \frac{N}{|\mathbf{B}|}(\mathbf{\Theta}(\mathbf{B})^\intercal \,(\dot{\mathbf{b}}-\mathbf{\Theta}(\mathbf{B})\,\tilde{\bm{\xi}})),
\label{eq:neg-log-post-grad}
\end{equation} where $\dot{\mathbf{b}} \in \mathbb{R}^{|\mathbf{B}|}$ denotes the time derivative of a single dimension of the system states, corresponding to the column vector $\tilde{\bm{\xi}} \in \mathbb{R}^m$.

For an appropriately chosen step size schedule $\eta_k$, it has been established that the Langevin MCMC converges to the target stationary distribution \citep{durmus2017nonasymptotic}. In practice, it may struggle to escape local traps when exploring multimodal distributions. To address these challenges, several methods, such as underdamped Langevin MCMC \citep{cheng2018underdamped}, Hamiltonian Monte Carlo \citep{neal2011mcmc}, cyclical SGMCMC \citep{zhang2020cyclical}, reSGLD \citep{chen2018accelerating}, etc., have been developed as extensions of the Langevin MCMC framework. These methods aim to improve computational efficiency and scalability, especially in high-dimensional settings. Related uncertainty-aware inference of Langevin dynamics has also been explored using Bayesian neural networks \citep{bae2025inferring}.

We consider reSGLD to balance exploration and exploitation when identifying system coefficients, which offers a blend of efficiency and robustness in traversing different energy levels to address large-scale non-convex sampling problems \citep{deng2020non,zheng2024constrained}. It employs a high-temperature chain for exploration and a low-temperature chain for exploitation. The sampling process is shown as \footnote{Although reSGLD can admit multiple temperatures, we utilize two chains for simplicity.}:
\begin{equation}
\begin{aligned}
\tilde{\bm{\xi}}_{k+1}^{(1)} & =\tilde{\bm{\xi}}_k^{(1)}-\eta_k \nabla  \tilde{L}(\tilde{\bm{\xi}}_k^{(1)}) + \sqrt{2\eta_k\tau_1} \bm{\zeta}_{k}^{(1)} \\
\tilde{\bm{\xi}}_{k+1}^{(2)} & =\tilde{\bm{\xi}}_k^{(2)}-\eta_k  \nabla\tilde{L}(\tilde{\bm{\xi}}_k^{(2)}) + \sqrt{2\eta_k\tau_2} \bm{\zeta}_{k}^{(2)},
\end{aligned}
\label{eq:resgld iteration}
\end{equation}

where $\tilde{\bm{\xi}}_{k}^{(1)}$ and $\tilde{\bm{\xi}}_k^{(2)}$ denote the sampling results of two chains with temperatures $\tau_1 < \tau_2$. Furthermore, we swap the Markov chains in (\ref{eq:resgld iteration}) with the corrected swapping function for the mini-batch setting, with the swap function defined as:
\begin{equation}
\tilde{S}\big(\tilde{\bm{\xi}}_k^{(1)}, \tilde{\bm{\xi}}_k^{(2)}\big) = e^{\left(1/{\tau_1} - 1/{\tau_2}\right) \left(\tilde{L}\big(\tilde{\bm{\xi}}_k^{(1)}\big) - \tilde{L}\big(\tilde{\bm{\xi}}_k^{(2)}\big) - \left(1/{\tau_1} - 1/{\tau_2}\right) \frac{\tilde{\sigma}^2}{C} \right) } 
\label{eq:swap rate}
\end{equation}
where $\tilde{\sigma}^2$ approximates the variance of $\tilde{L}\big(\tilde{\bm{\xi}}_k^{(1)}\big) - \tilde{L}\big(\tilde{\bm{\xi}}_k^{(2)}\big)$ and $C$ acts as an adjustment to balance acceleration and bias. The algorithm proceeds iteratively until a specified number of iterations is reached. Generate uniform $u\in[0,1]$, if $u<\tilde{S}$, then perform the swapping between two chains. The parameter sets $\{\tilde{\bm{\xi}}_k^{(1)}\}_{k=1}^{K+1}$ are produced as outputs for analytical purposes.

While exact Bayesian inference can be computationally demanding, reSGLD leverages minibatch gradients and parallel chains to efficiently approximate the posterior distribution. This strategy avoids the restrictive assumptions of conjugate-prior Bayesian methods \citep{fung2024} and provides richer uncertainty information, including multimodality and credible intervals.

\subsection{
Active Learning for Data-Efficient System Discovery
}
The integration of AL into data-driven discovery of dynamical systems aims to address two fundamental challenges in learning parsimonious dynamical models from limited and costly data.
First, while accurate state derivatives $\mathbf{\dot{U}}$  are the cornerstone of the regression in Eq. \eqref{eq:system}, their acquisition is often a very expensive component of the discovery workflow. High-quality estimation requires dense, clean data, but collecting such data uniformly across the state space is highly inefficient and often infeasible. This highlights the critical need for a targeted data acquisition strategy with AL.
Second, dynamical systems often exhibit heterogeneous information content across their state space. Sparse governing terms dominate in certain regimes (e.g., near bifurcations or transient states), while other regions contribute minimally to identifying the dynamical systems. 
Passive sampling strategies are inefficient and often fail to collect sufficient data in the information-rich regions. This oversight can lead to models that are not just inaccurate but fundamentally misrepresent the true system dynamics.

AL directly targets these issues by prioritizing data acquisition, where the uncertainty in the learned dynamics is maximized. By tackling \eqref{eq: linear} within a Bayesian framework, posterior uncertainties over the coefficients $\mathbf{\Xi}$ in \eqref{eq: linear} can be quantified. Regions of high predictive entropy or variance in $\mathbf{\dot{U}}$ correspond to states in which the model cannot confidently distinguish between candidate terms in the library $\mathbf{\Theta}(\mathbf{U})$. Actively querying derivatives in these states maximizes the information gain on $\mathbf{\Xi}$, which efficiently resolves ambiguities in the sparsity pattern \citep{riis2023bayesian, pickering2022discovering, Fasel_2022}. This method, which acquires data with the highest predictive uncertainty, is also known as Active Learning MacKay (ALM) \citep{MacKay1992, gramacy2009adaptive}.

To formulate AL, we first denote the pool of potential design points as $\mathcal{D}=\{\mathbf{U} ,\dot{\mathbf{U}}\}$, and denote the set of selected points as $\mathcal{M}$. With the information of the posterior $p(\mathbf{\Xi}|\mathbf{U})$, we sample $P$ groups of the model parameters from the posterior distribution and calculate $\dot{\mathbf{u}}^{pred}_{j}(t_i)$ for each model $j$ at time $t_i$, with the mean of the prediction denoted as $\bar{\dot{\mathbf{u}}}^{pred}(t_i)$. The predictive variance is calculated as 
\[
\sigma^2({\dot{\mathbf{u}}}(t_i))=\frac{1}{P}\sum_{i=1}^{P}(\dot{\mathbf{u}}^{pred}_{j}(t_i)-\bar{\dot{\mathbf{u}}}^{pred}(t_i))^2,
\]
leading to the selection of potential design points with high uncertainty. In this paper, $\sigma^2(\cdot)$ is referred to as the uncertainty acquisition function.
A key limitation observed in our experiments is that using the uncertainty acquisition function alone can lead to the redundant selection of points clustered in a single region. This occurs because while a query point may be informative, the uncertainty in its neighborhood often remains high, attracting subsequent samples. These clustered points offer diminishing returns in terms of new information about the overall system, resulting in a model that overfits the dynamics of one region while failing to learn from other potentially critical areas.

To address this clustering issue, BLADE adopts a hybrid acquisition strategy that balances uncertainty-driven exploitation with space-filling exploration. We augment the uncertainty measure with a density-adjusted maximin distance criterion \citep{loeppky2010,yu2010}. This criterion promotes broader coverage by selecting new points that maximize the minimum distance to the existing dataset.
The space-filling criterion is adjusted by the density of the data $\mathbf{u}(t_i)$, and is defined as:
 \begin{equation}
     d(\mathbf{\Theta}(\mathbf{u}(t_i)))=\text{min} ||\mathbf{\Theta}(\mathbf{u}(t_i))-\mathbf{\Theta}(\mathbf{u}(t_j))||_2 \cdot \text{density}(\mathbf{u}(t_i))^{\lambda}, \text{for all} ~\mathbf{u}(t_j) \in \mathcal{M},j\neq i,
 \end{equation} 
 where the term $\lambda \geq 0$ is a hyperparameter that controls the strength of the density penalty applied to the distance measure. In the limiting case where $\lambda=0$, the penalty vanishes, and the criterion reverts to the original maximin design. A value of $\lambda=0.5$, for instance, balances the push for exploration (distance) with a preference for sampling in already populated regions (density). The objective of density adjustment is to penalize low-density points because they are more likely to have a large maximin distance, especially if they also have a large magnitude.
Without this density adjustment, the search for space-filling points would be biased towards high-magnitude outliers and disregard potentially more informative points in dense clusters. This correction ensures the criterion promotes exploration across representative regions rather than simply selecting isolated points, and it also prevents overlap with the uncertainty-based component of our acquisition function. 
 
Density-based adjustment has been adopted in AL research for both classification and regression tasks to achieve different goals, such as increasing the representativeness of the data \citep{wang2021density}, mitigating the sensitivity of the Euclidean distance to the scale and distribution of the data \citep{donmez2007paired}, and preventing outliers added to the training data \citep{zhu2008active}. 
Several methods can be used to estimate data density, including Kernel Density Estimation (KDE), cosine distance-based metrics, and K-Nearest Neighbors (KNN). For this work, we select the KNN-based approach due to its advantages over the alternatives, which is calculated as the inverse of one point's average distance to its K nearest neighbor. Compared to the computationally intensive KDE, KNN offers computational simplicity and directly captures local data density without evaluating kernel functions over all data. Furthermore, unlike cosine similarity, which captures only vector direction, KNN preserves both magnitude and directional information, which is crucial for identifying representative samples in our feature space.

The final acquisition function is constructed by combining the uncertainty and space-filling criteria. To ensure that they are on a comparable scale, both criteria are first standardized to the range $[0, 1]$, yielding the normalized uncertainty $\tilde{\sigma}^2({\dot{\mathbf{u}}(t_i)})$ and the normalized distance $\tilde{d}(\mathbf{\Theta}(\mathbf{u}(t_i)))$. The final acquisition score is calculated as a weighted sum:
 \begin{equation}
     C(\mathbf{u}(t_i)) = \alpha \,\tilde{\sigma}^2({\dot{\mathbf{u}}}(t_i)) + (1-\alpha) \,\tilde{d}(\mathbf{\Theta}(\mathbf{u}(t_i))),
 \end{equation}
where a tunable hyperparameter $\alpha \in [0, 1]$ balances their relative contributions to the final acquisition score.

\begin{algorithm}[!htbp]
\SetAlgoLined
\KwIn{Pool of potential design points $\mathcal{D}$; number of initial points $n$; batch size of selected points per round $m$, maximum allowed sample size $N_{max}$, tolerance of convergence $\textit{Tol}$.}
\KwOut{Stabilized model with coefficients $\mathbf{\Xi}$}
Randomly select $n$ points, $\mathbf{u}(t_1), \dots, \mathbf{u}(t_n) \in \mathbf{U}$ with derivatives $\dot{\mathbf{u}}(t_1), \dots, \dot{\mathbf{u}}(t_n)$\ available;\\
Set of selected points $\mathcal{M}=\{(\mathbf{u}(t_1),\dot{\mathbf{u}}(t_1)),\dots,(\mathbf{u}(t_n),\dot{\mathbf{u}}(t_n))\}$, $\mathcal{D} \leftarrow \mathcal{D}/\mathcal{M}$ \tcp*[f]{i.e., exclude selected data from the design pool};\\
Initialize $\mathcal{E}_0$ as a very large number;

\While{$|\mathcal{M}|\leq N_{max}$}{
Use $\mathcal{M}$ points to construct the library $\mathbf{\Theta}$ and perform BLADE to estimate the posterior distribution $p(\boldsymbol{\mathbf{\Xi}} \mid \mathcal{M})$\;
    \For{all potential design points $\mathbf{u} \in \mathcal{D}$}{
        Calculate $C(\mathbf{u})$
    }
    \For{$i = 1, \dots, m$}{Select the sample $\mathbf{u}$ with highest $C(\mathbf{u})$;\\
    Collect additional data and compute derivative $\mathbf{\dot{u}}$ if not available;\\
    $\mathcal{M} \leftarrow \mathcal{M} \cup {(\mathbf{u},\dot{\mathbf{u}})},\mathcal{D} \leftarrow \mathcal{D}/\mathcal{M} $
    }
    Calculate $\mathcal{E}$;\\
    \If{$\frac{|\mathcal{E}-\mathcal{E}_0|}{\mathcal{E}}< \text{Tol}$}{break;}
    $\mathcal{E}_0 \leftarrow \mathcal{E}$;

}
\caption{Active Learning in BLADE: Iteratively improve the model by selecting design points that maximize the acquisition function, updating the posterior, and retraining the model until convergence.}
\label{al:activelearning}
\end{algorithm}

\subsection{The Proposed Algorithm}

\begin{figure}
    \centering
    \includegraphics[width=0.8\linewidth]{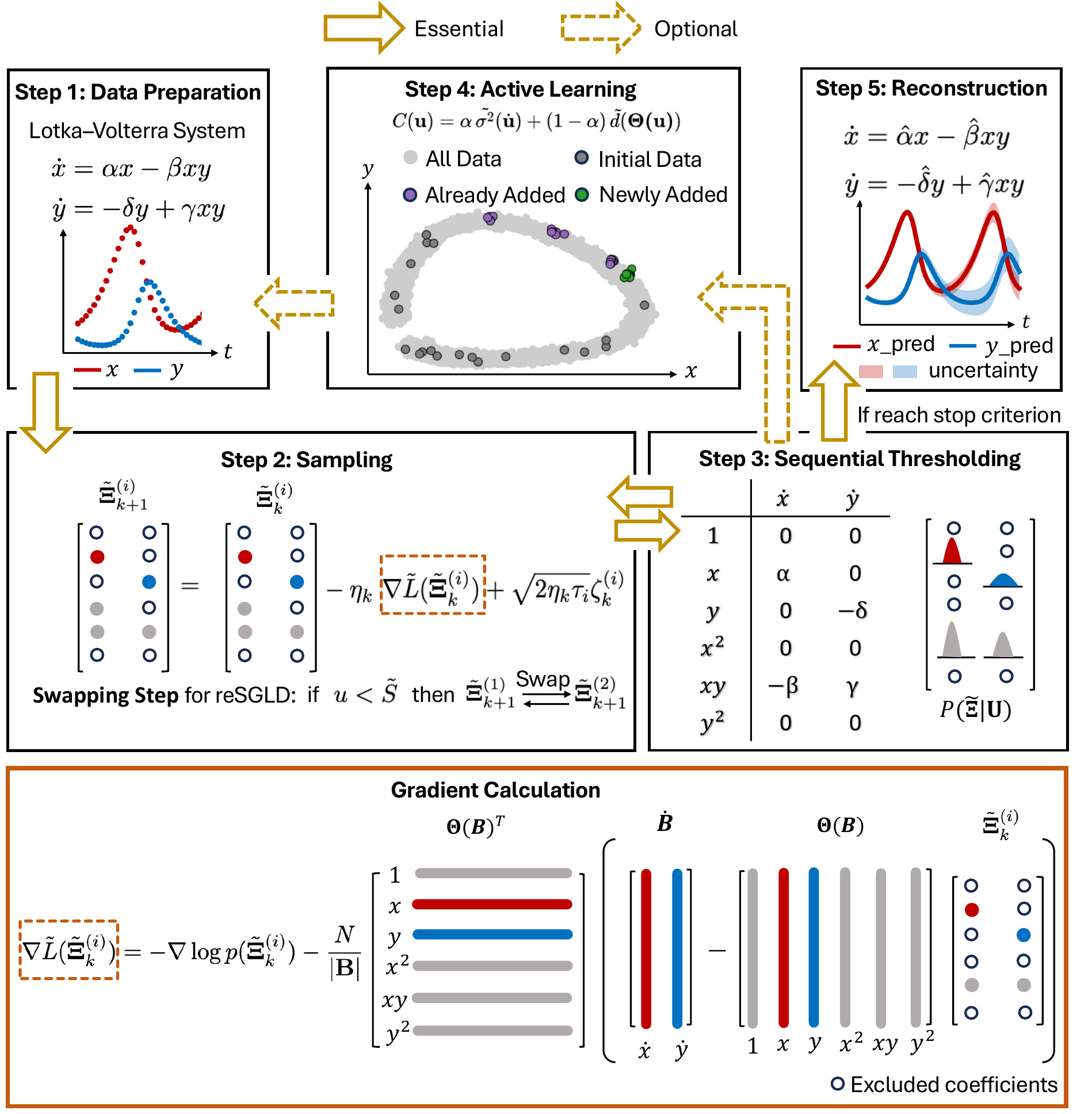}
    \caption{The BLADE framework is demonstrated through the identification of the Lotka-Volterra system. The process begins with \textbf{Step 1: Data Preparation}, where the dataset $\mathbf{U}$ is obtained from the differential equations under investigation. \textbf{Step 2: Sampling:} The (stochastic) gradient $\tilde{L}(\tilde{\mathbf{\Xi}}_k)$ is computed as outlined in the \textbf{Gradient Calculation} procedure (bottom). \textbf{Step 3: Sequential Thresholding:} Based on the sampling outcome, thresholding is applied by removing bases with small absolute coefficients from the library; for example, as in the plot, the $xy$ basis for state $x$ is excluded by thresholding. If the retained library differs from the previous iteration, another sampling round is conducted. \textbf{Step 4: Active Learning:} If AL is adopted, continue to add new training points according to the hybrid acquisition function and restart the training. \textbf{Step 5 Reconstruction:} Once the final posterior distribution $P(\tilde{\mathbf{\Xi}}|\mathbf{U})$ is obtained, the system is reconstructed with uncertainty quantification.}
    \label{fig:blade}
\end{figure}

We now introduce the procedure for implementing BLADE. This framework can be applied to scenarios with sufficient data, whether clean or noisy, utilizing reSGLD for parameter sampling, and it can also be applied to scenarios with limited data, where AL will help to minimize data acquisition cost. The proposed BLADE framework is summarized in Figure \ref{fig:blade}.

\textbf{Data preparation.}
The first step requires constructing the time derivatives, $\dot{\mathbf{U}}$, and the library of candidate functions, $\mathbf{\Theta}(\mathbf{U})$. When $\dot{\mathbf{U}}$ is not directly measured, it is numerically approximated from the collected data $\mathbf{U}$. We employ the Finite Difference method for this task due to its computational efficiency and the low-noise characteristics of our dataset. The composition of the library $\mathbf{\Theta}(\mathbf{U})$ depends on the type of system. For ODEs, the library is composed of polynomial functions of the system's state variables. For PDEs, the library is constructed after computing the necessary spatial partial derivatives from the data, for which the Finite Difference method is also adopted.

\textbf{Parameter Sampling.}
Our methodology for estimating the posterior distribution of model parameters is based on a Markov Chain Monte Carlo (MCMC) sampling process. The process is initialized by defining priors for the model coefficients. The target distribution is specified by treating the negative log-posterior, $\tilde{L}(\mathbf{\Xi})$ in \eqref{eq:neg-log-post-est}, as the energy function. To ensure the chain properly explores the target posterior, we set the temperature $\tau=1$ and incorporate a Metropolis-Hastings (MH) acceptance criterion to maintain detailed balance. An initial warm-up phase is run to guarantee the sampler has reached its stationary distribution, and these preliminary samples are discarded. The final approximation of the posterior is then constructed from all samples collected post-warm-up.
Subsequently, we achieve a sparse result by applying sequential thresholding to the posterior coefficient samples from Eq.\eqref{eq:resgld iteration}. In this step, any coefficient with a posterior mode smaller than a given threshold $c$ is removed from the candidate basis, as proposed in \citep{brunton2016discovering}. After the coefficient support stabilizes, the Langevin samples quantify coefficient uncertainty within the selected model class. Accordingly, our UQ for coefficients is best interpreted as conditional on the discovered support, not as Bayesian uncertainty that marginalizes over model structure.


\textbf{Data Acquisition for Active Learning.}
In the context of data scarcity, AL is utilized to augment the set of selected points, $\mathcal{M}$, with new potential design points. The implementation of this AL procedure is formally presented in Algorithm~\ref{al:activelearning}. After new points are added, the \textbf{data preparation} step is repeated to update the derivative estimates and the basis library with the expanded set $\mathcal{M}$. Upon the completion of model training, the process advances to the reconstruction of the model and the subsequent estimation of its uncertainty.

\textbf{Model Reconstruction and Uncertainty Quantification.}
For both ODE and PDE systems, we generate predictions and quantify uncertainty by leveraging the posterior distribution of the coefficients, $\tilde{\mathbf{\Xi}}$. For ODEs, predicting the system's evolution requires solving the integral:
\[
\hat{\mathbf{x}}^\top(t; \tilde{\mathbf{\Xi}}, \mathbf{x}_0) = \mathbf{x}_0^\top + \int_{t_0}^t \mathbf{\Theta}(\mathbf{x}(t')) \tilde{\mathbf{\Xi}} \, dt'.
\]
As this is typically analytically intractable, we instead create an ensemble of predicted trajectories by numerically integrating the system for each posterior sample of $\tilde{\mathbf{\Xi}}$. This ensemble-based approach provides both a mean prediction and a measure of trajectory uncertainty arising from the variance in the coefficients. For PDEs, trajectories are similarly generated from each coefficient sample, using either numerical solvers (e.g., via Fast Fourier Transform) or analytical solutions where available. The uncertainty in the model coefficients themselves is directly captured by the distribution of the posterior samples.

\section{
System Discovery with Uncertainty Quantification
}\label{sec:model_discovery}
\label{section:experiment}
To assess the performance of BLADE under conditions of ample data availability, four experiments were performed without the use of AL: two for ODE systems (Lotka-Volterra and Lorenz) and two for PDE systems (Burgers' equation and convection–diffusion equation). The baseline models used are SINDy (or PDE-FIND for PDEs) and UQ-SINDy. SINDy, a frequentist method, is less robust to noise compared to BLADE, as we will demonstrate later. UQ-SINDy, a Bayesian approach, provides UQ and identifies sparse systems. 
However, its reliance on sparsity-inducing priors introduces sensitivity to prior selection, which can hinder accurate recovery of the true system when assumptions about sparsity are misaligned with the underlying dynamics.
Additionally, we compare different Langevin MCMC methods within BLADE, including SGLD, cyclical SGMCMC, and reSGLD.

Cyclical SGMCMC follows the same iteration formula as (\ref{eq:ULA}), but the step size is adjusted every round to balance between exploration and sampling. The step-size at iteration $k$ is defined as
\begin{equation*}
    \eta_k = \frac{\eta_0}{2} \left[ \cos \left( \pi \frac{\text{mod}(k-1, \lfloor K/M \rfloor)}{\lfloor K/M \rfloor} \right) + 1 \right],
\end{equation*}
where $\eta_0$ is the initial step-size, $M$ is the number of cycles, and $K$ is the number of total iterations. During the implementation of Langevin MCMC methods, the Metropolis-Hastings step is performed to ensure that the distribution of the iterated instances $\tilde{\mathbf{\Xi}}_{k}$ sampled in round $k$ converges to the correct distribution $\Pi$ when $k \rightarrow \infty$ \citep{dwivedi2019log,chewi2021optimal}.
Unless otherwise specified, the default sampling method adopted in this work is the 2-chain reSGLD, and the prior adopted is the regularized horseshoe with $\nu=4,s=2$ as in \citep{hirsh2022sparsifying}. For reSGLD, the ratio \(\tilde{\sigma}^2/C\) in Eq.~\eqref{eq:swap rate} is treated as a single effective hyperparameter as a common practice and is tuned during training. Empirically, good performance is obtained when the post--warm-up swapping rate between two chains stays around \(5\%\)–\(15\%\). In our experiments, we set the temperature of two chains as \(\tau_1=1\) and \(\tau_2=2\), and tune \(\tilde{\sigma}^2/C\) accordingly to reach this target range.

Before conducting the experiments, we outlined the setup details. The noise-free data of ODEs is simulated numerically with solve\_ivp method from the Scipy package in Python using the RK45 integration method, setting $10^{-3}$ for relative tolerance and $10^{-6}$ for absolute tolerance. The 1D Burgers' system was converted into a system of ODEs using Fast Fourier Transform, and the odeint function from scipy.integrate was used to numerically generate noise free data for this system. The convection–diffusion system was solved analytically in Python with Numpy. The noisy data is generated by adding independent and identically distributed (i.i.d.) Gaussian white noise with zero mean is added to the clean data. The variance of the noise for each dimension of the state is calculated as $\sigma = p \cdot \text{std}(\mathbf{U}_i)$, where $\text{std}(\mathbf{U}_i)$ is the standard deviation of the clean data along that dimension $i$, and $p$ is the chosen noise level (e.g., 5\%). All models are trained with the same derivative $\dot{\mathbf{U}}$ and library $\mathbf{\Theta}(\mathbf{U})$ prepared, and the same threshold if sequential thresholding is adopted, making sure a fair comparison. We note that the noise levels used in the ODE and PDE benchmarks are not uniform. In particular, the ODE experiments are conducted with 5\% additive noise, while the PDE experiments use lower noise levels (0.2\% for Burgers’ equation and 0.1\% for convection--diffusion). This reflects a practical limitation of the current pipeline: for PDE discovery, numerical derivative estimation and candidate-library construction are substantially more sensitive to measurement noise. Therefore, the PDE results in this section should be interpreted as validation in low-noise regimes rather than a comprehensive stress test of high-noise robustness.

We used three error metrics to evaluate the performance of different models, namely, Total Error Bar, Mean Squared Error (MSE) and Akaike Information Criterion (AIC). Total Error Bar is uncertainty-aware and works well in the Bayesian setting.
The Total Error Bar criterion is defined as below following \cite{zhang2018robust}:
\begin{equation}
    \mathcal{E}(\tilde{\mathbf{\Xi}}) = \sum_{\substack{i=1 \\ \tilde{\Xi}_i \neq 0}}^{md} \frac{\tilde{s}_{i}^2}{\tilde{\Xi}_i^2},
\label{eq:error-bar}
\end{equation}
with $md$ being the total number of entries in the parameter.
Here, \(\tilde{\Xi}_i\) denotes one entry of the predicted posterior mode of the coefficients and \(\tilde{s}_i\) its posterior standard deviation; thus, Total Error Bar is a squared coefficient of variation (CV)-type aggregate uncertainty metric over the retained coefficients. Coefficients thresholded to zero are excluded to avoid numerical instability from near-zero denominators.
The Bayesian Total Error Bar works both in model comparison and in threshold selection. The example of threshold selection is given in Section \ref{sec:4.4}. A lower Total Error Bar indicates a more reliable coefficient result. MSE quantifies the discrepancy between the predicted derivatives of the identified model states and the actual measurements, which is calculated as
\begin{equation*}
\text{MSE} = \frac{1}{N d} \sum_{i=1}^{N} \sum_{j=1}^{d} (\dot{\mathbf{U}}_{ij} - \tilde{\dot{\mathbf{U}}}_{ij})^2,
\end{equation*}
where $\dot{\mathbf{U}}$ denotes the true derivative (noise-free) and $\tilde{\dot{\mathbf{U}}}$ denotes the predicted derivative, $N$ and $d$ representing the number of collection and state dimension, respectively. The smaller the MSE, the better the model learned fits the data.
AIC \cite{akaike1974new, schwarz1978estimating} deals with the trade-off between the goodness of fit of the model and the simplicity of the model. Here, AIC can be calculated based on the MSE,
\[
    \text{AIC} = 2k + Nd\, \ln(\text{MSE}),
\]
where $k$ denotes the number of non-zero coefficients. The smaller the AIC, the more favorable the model, as it effectively balances model fit and parsimony, penalizing unnecessary complexity while rewarding accuracy. Furthermore, we assess uncertainty calibration using the Prediction Interval Coverage Probability (PICP), defined as the fraction of ground-truth values that fall within the 
posterior confidence interval computed from predictive samples. In addition, we report the Mean Prediction Interval Width (MPIW), defined as the average width of the same posterior confidence interval, which complements PICP by quantifying interval sharpness.

\subsection{Lotka-Volterra System}
\label{lotka_volterra}
We begin by assessing the method using the Lotka-Volterra system, often known as the predator-prey model. This system is extensively utilized to depict the interactions between two competing populations. Initially formulated by Lotka to simulate chemical reactions, it has since become a foundational framework for analyzing dynamics in biological systems and economic models. The Lotka-Volterra system can be characterized by the following nonlinear ODE system:
\begin{equation}
\begin{cases}
\dot{x} = \alpha x - \beta xy, \\
\dot{y} = -\delta y + \gamma xy.
\end{cases}
\end{equation}

In this experiment, we simulate the data with the initial condition $[x_0,y_0]=[10,5]$, and the system parameters $\alpha = 1.0,
\beta = 0.1,
\delta = 1.5,
\gamma = 0.075$. Starting from $t=0$, $5,000$ time steps of states are simulated, with $\Delta t=5\cdot 10^{-3}$. The library consists of polynomial terms up to order 2, with the library basis being $[1,x(t),y(t),x(t)^2,x(t)y(t),y(t)^2]$. The collected data are then corrupted with 5\% additive noise. For the sequential thresholding steps, the threshold is set to 0.065, which is appropriately chosen to preserve the smallest true system parameter ($\gamma=0.075$). 

The coefficients learned from various models are presented in Table \ref{lotka_table}. The results indicate that BLADE performs well in accurately identifying the true system structure, with no extraneous or missing terms, whereas SINDy and UQ-SINDy introduce additional terms not present in the true system. BLADE shows better performance compared to other baseline methods, as reflected in lower MSE and AIC values. 

\begin{figure}[h]
    \centering
    \includegraphics[width=1\linewidth]{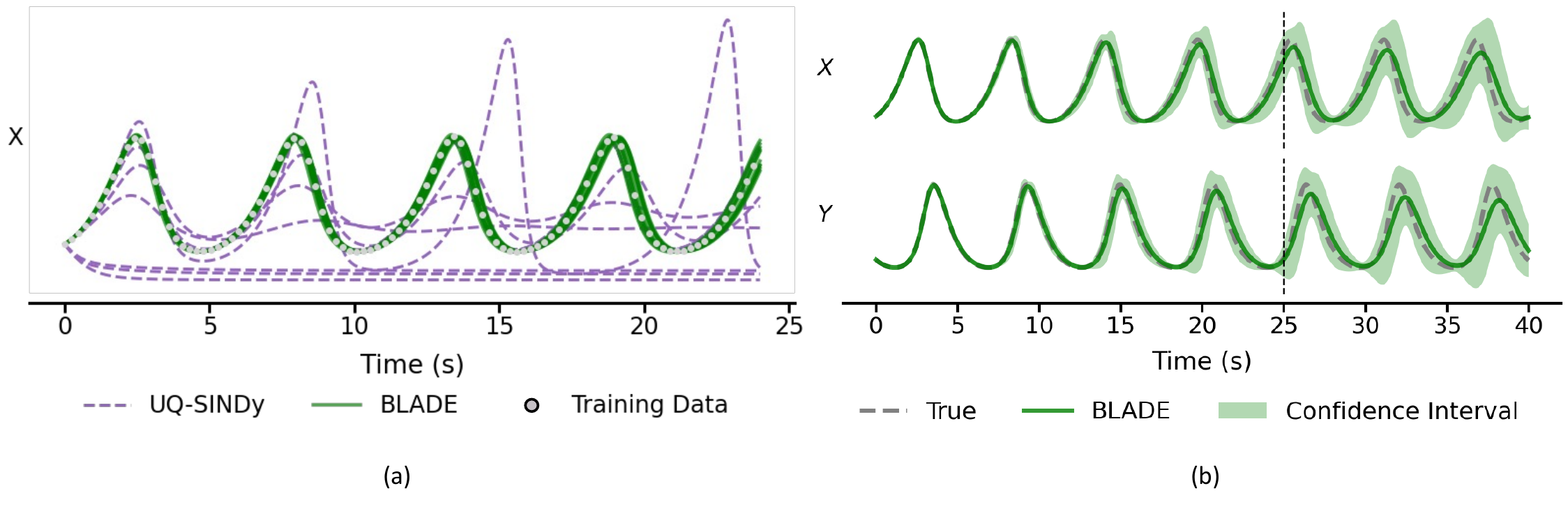}
    \caption{(a) Predicted trajectories of UQ-SINDy and BLADE on the Lotka-Volterra system. Gray dots represent training data, purple dashed lines are trajectories generated by UQ-SINDy, and green solid lines are yielded by BLADE. Both UQ-SINDy and BLADE run 10 times with different prior initializations. 
    (b) Trajectory prediction and confidence interval with BLADE on the Lotka-Volterra system, with the 95\% confidence interval. The dashed vertical line marks the transition from in-sample prediction to out-of-sample prediction. }
    \label{fig:lotka_compare}
\end{figure}

Figure \ref{fig:lotka_compare}(a) illustrates the in-sample predictions using coefficients learned from 10 different groups of priors with varying coefficients for the horseshoe prior. With $\tau_0 \sim U(0.05,0.3)$, $\nu \sim U(0,5)$, $s\sim U(0,3)$, where $U(\cdot,\cdot)$ represents uniform distribution. 
The plot reveals significant variability in UQ-SINDy predictions, with certain groups exhibiting discrepancies in tracking the state fluctuations. This robustness is mainly due to the fact that BLADE's coefficient updates are primarily driven by likelihood rather than the specific choice of prior, reducing sensitivity to prior selection. Additionally, the use of sequential thresholding helps achieve system sparsity rather than relying merely on sparsity-inducing priors. UQ-SINDy, which also uses a Bayesian approach to estimate the posterior distribution of the coefficients, relies on sparsity-inducing priors such as the horseshoe prior to exclude extraneous terms. The performance of UQ-SINDy is sensitive to the choice of prior, and in some cases, may not yield the desired level of system sparsity.

Figure \ref{fig:lotka_compare}(b) provides 95\% confidence intervals in the prediction space. The uncertainty coverage effectively captures the true trajectory of the Lotka-Volterra system, particularly in cases where the model's predictions deviate from the truth. When the model predictions align well with the true trajectory, the uncertainty coverage remains appropriately modest. For the Lotka--Volterra system, the 95\% prediction intervals produced by the model trained on noisy data achieve a PICP of 100\%. The corresponding MPIW values are 6.25 for the in-sample intervals and 16.75 for the out-of-sample intervals. Relative to the mean ground-truth prediction magnitudes of 57.85 and 60.29, respectively, these interval widths suggest that the perfect coverage is not obtained by excessively wide intervals. This highlights the reliability of the BLADE in both accurate prediction and uncertainty estimation.

\begin{table}[h]
\centering
\scriptsize 
\caption{Learned coefficients with different models on the Lotka-Volterra System with 5\% noise}
\adjustbox{center=\textwidth}{
\begin{tabular}{lcccccccc}
\toprule
\multirow{2}{*}{Basis} & \multicolumn{2}{c}{True} & \multicolumn{2}{c}{SINDy} & \multicolumn{2}{c}{UQ-SINDy} & \multicolumn{2}{c}{BLADE} \\
\cmidrule(lr){2-3} \cmidrule(lr){4-5} \cmidrule(lr){6-7} \cmidrule(lr){8-9}
 & X & Y & X & Y & X & Y & X & Y \\
\midrule
1 & - & - & 4.158e-1 & - & 1.814e-2 & -4.512e-2 & - & - \\
x & 1.000e0 & - & 9.858e-1 & - & 9.613e-1 & -2.194e-2 & \textbf{9.966e-1} & - \\
y & - & -1.500e0 & - & -1.428e0 & 1.793e-2 & -1.301e0 & - & \textbf{-1.469e0} \\
$x^2$ & - & - & - & - & 3.281e-4 & 9.838e-4 & - & - \\
$xy$ & -1.000e-1 & 7.500e-2 & -1.000e-1 & \textbf{7.446e-2} & -9.707e-2 & 6.935e-2 & \textbf{-1.000e-1} & 7.386e-2 \\
$y^2$ & - & - & - & - & -1.957e-3 & -4.288e-3 & - & - \\
\midrule
Total Error Bar & \multicolumn{2}{c}{-} & \multicolumn{2}{c}{-} & \multicolumn{2}{c}{1.017e3} & \multicolumn{2}{c}{\textbf{1.747e-3}} \\
\midrule
MSE & \multicolumn{2}{c}{-} & \multicolumn{2}{c}{1.248e1} & \multicolumn{2}{c}{4.470e1} & \multicolumn{2}{c}{\textbf{7.977e0}} \\
\midrule
AIC & \multicolumn{2}{c}{-} & \multicolumn{2}{c}{2.525e4} & \multicolumn{2}{c}{3.802e4} & \multicolumn{2}{c}{\textbf{2.077e4}} \\
\bottomrule
\label{lotka_table}
\end{tabular}}
\end{table}

\subsection{Lorenz System}

Consider the Lorenz system of the form:
\begin{equation}
\begin{cases}
    \dot{x} = \sigma (y - x), \\
    \dot{y} = x (\rho - z) - y, \\
    \dot{z} = xy - \beta z,
\end{cases}
\label{eq:lorenz_system}
\end{equation}
with the system dynamics being governed by three parameters: the Prandtl number $\sigma$, the Rayleigh number $\rho$, and the aspect ratio $\beta$. This system is renowned for its chaotic behavior and is widely used to study complex, nonlinear dynamics in fields such as meteorology, fluid dynamics, and climate modeling. The interplay between these parameters gives rise to intricate and often unpredictable trajectories, making the Lorenz system a classic example of deterministic chaos. We simulate the system with initial conditions $\begin{bmatrix}
    x(0),y(0),z(0)\end{bmatrix}=\begin{bmatrix}
        -8,8,27
    \end{bmatrix}$, starting from $t=0$ and end at $t=5.5$ with time intervals $\Delta t = 1\times 10^{-3}$.
We consider the library consisting of polynomial terms up to the second order, which consists of $10$ basis: $\begin{bmatrix}
1,x(t), y(t) ,z(t) , x^2(t),x(t)y(t) , x(t)z(t), y^2(t), y(t)z(t), z^2(t)
\end{bmatrix}$. The training data is injected with 5\% additive noise.

The result of the BLADE experiment is compared with SINDy and UQ-SINDy in Table \ref{lorenz_table}. The thresholds here for SINDy and BLADE are both set to 0.5. As shown in the table, BLADE successfully identified all the correct terms to be included in the Lorenz system, with no missing or redundant terms. However, SINDy has redundant terms in the identification of $Y$ and $Z$. In this example, SINDy faces challenges in accurately learning the system even by choosing the threshold that favors the magnitude of the coefficients learned. For example, increasing the threshold to 0.6 will exclude the constant terms in SINDy for $Y$ and $Z$, but the learned coefficient would result in an increase in MSE (4.503) and AIC (4.489$\times 10^{3}$); increasing the threshold to 0.7 would cause a missing basis $y$ for state $Y$. The presence of many redundant bases in the model identified by UQ-SINDy highlights a potential limitation of relying solely on sparse-inducing priors. In contrast, our results indicate that the BLADE method yields more favorable outcomes across the three considered criteria.

\begin{figure}
    \centering
    \includegraphics[width=1\linewidth]{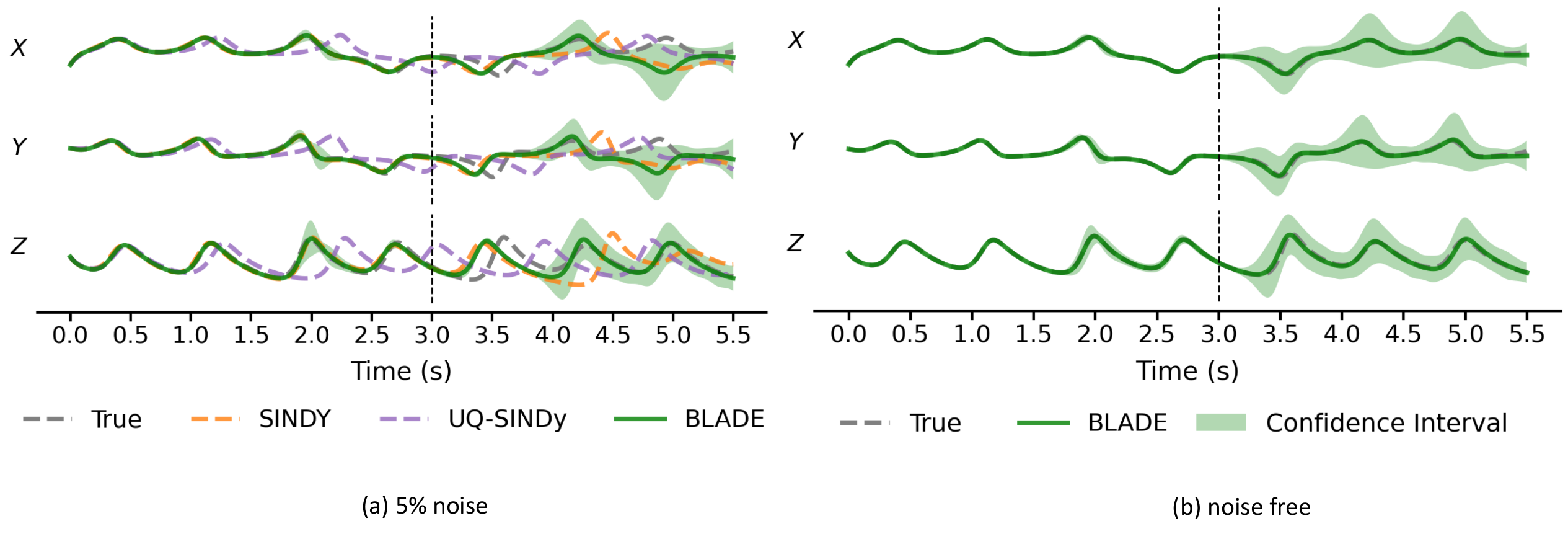}
    \caption{(a) Trajectory prediction and confidence interval with SINDY, UQ-SINDy, and BLADE on the Lorenz system. Training data is injected with 5\% noise. The dashed vertical line marks the transition from in-sample prediction to out-of-sample prediction. The confidence interval (95\%) for BLADE is illustrated as a green shaded area.  (b) Trajectory prediction and confidence interval with BLADE on the Lorenz system. Training data is noise-free. }
    \label{fig:lorenz-uq}
\end{figure}

As shown in Fig.~\ref{fig:lorenz-uq}, when the model is trained on noisy data, the 95\% prediction intervals achieve a PICP of 92\% for in-sample prediction and 67\% for out-of-sample prediction. The corresponding MPIW values are 0.80 for the in-sample intervals and 3.51 for the out-of-sample intervals. Relative to the mean ground-truth prediction magnitudes of 10.09 and 9.14, respectively, these interval widths indicate that the reported coverage is not achieved through trivially inflated intervals. 
By contrast, in panel (b), when the data are noise-free, BLADE exhibits stronger UQ performance, with the PICP of the 95\% prediction intervals reaching 100\%. The corresponding MPIW values are 0.37 for the in-sample intervals and 4.67 for the out-of-sample intervals. These results suggest that uncertainty quantification is substantially more reliable in the noise-free setting, whereas under high noise the prediction intervals become less well calibrated, especially for longer-horizon out-of-sample prediction. This observation also indicates that, in noisy settings, limiting the prediction horizon or periodically reinitializing from newly collected observations may help mitigate the accumulation of predictive error.

A key difference between BLADE and UQ-SINDy is the choice of sampler: BLADE uses reSGLD, whereas UQ-SINDy uses the No-U-Turn Sampler (NUTS). Since one NUTS transition and one reSGLD update are not directly comparable in terms of either statistical information content or computational structure, we use steps/s only as a measure of raw computational throughput under matched hardware, warm-up, and sampling-step budgets. To this end, we benchmark both samplers on the noise-free Lorenz dataset under identical settings. Under this benchmark, post-warm-up reSGLD sustained 25.29 steps/s, whereas NUTS sustained 1.14 steps/s. This highlights a substantial practical difference in computational throughput between the two samplers under matched implementation settings.

\begin{table}[h]
\centering 
\scriptsize
\caption{Learned coefficients with different models on the Lorenz System with 5\% noise.}
\adjustbox{center=\textwidth}{
\setlength{\tabcolsep}{1.5pt} 
\begin{tabular}{@{}l@{}cccccccccccc}
\toprule
\multirow{2}{*}{Basis} & \multicolumn{3}{c}{True} & \multicolumn{3}{c}{SINDy} & \multicolumn{3}{c}{UQ-SINDy} & \multicolumn{3}{c}{BLADE} \\
\cmidrule(lr){2-4} \cmidrule(lr){5-7} \cmidrule(lr){8-10} \cmidrule(lr){11-13}
 & X & Y & Z & X & Y & Z & X & Y & Z & X & Y & Z \\
\midrule
1 & - & - & - & - & 5.672e-1 & -5.545e-1 & 1.855e-2 & 4.449e-2 & -2.583e-1 & - & - & - \\
x & -1.000e1 & 2.800e1 & - & \textbf{-1.016e1} & 2.714e1 & - & -1.096e-1 & 2.250e1 & -3.965e-3 & -9.942e0 & \textbf{2.797e1} & - \\
y & 1.000e1 & -1.000e0 & - & \textbf{1.014e1} & -6.566e-1 & - & 7.420e-1 & 5.070e-1 & -3.649e-3 & 9.981e0 & \textbf{-9.111e-1} & - \\
z & - & - & -2.667e0 & - & - & \textbf{-2.655e0} & 2.712e-2 & 8.359e-3 & -1.377e0 & - & - & -2.681e0 \\
$x^2$ & - & - & - & - & - & - & -1.728e-3 & -6.264e-3 & 1.707e-1 & - & - & - \\
$xy$ & - & - & 1.000e0 & - & - & \textbf{1.001e0} & -2.500e-2 & -4.793e-4 & 8.803e-1 & - & - & \textbf{1.001e0} \\
$xz$ & - & -1.000e0 & - & - & -9.856e-1 & - & -3.124e-1 & -8.601e-1 & 2.113e-3 & - & \textbf{-1.007e0} & - \\
$y^2$ & - & - & - & - & - & - & 1.337e-2 & 3.945e-3 & -1.076e-2 & - & - & - \\
$yz$ & - & - & - & - & - & - & 3.202e-1 & -1.363e-2 & -5.477e-4 & - & - & - \\
$z^2$ & - & - & - & - & - & - & 1.384e-3 & 3.450e-4 & -4.770e-2 & - & - & - \\
\midrule
Total Error Bar & \multicolumn{3}{c}{-} & \multicolumn{3}{c}{-} & \multicolumn{3}{c}{5.267e4} & \multicolumn{3}{c}{\textbf{1.435e-3}} \\
\midrule
MSE & \multicolumn{3}{c}{-} & \multicolumn{3}{c}{3.913e2} & \multicolumn{3}{c}{6.623e3} & \multicolumn{3}{c}{\textbf{1.564e2}} \\
\midrule
AIC & \multicolumn{3}{c}{-} & \multicolumn{3}{c}{5.374e4} & \multicolumn{3}{c}{7.924e4} & \multicolumn{3}{c}{\textbf{4.549e4}} \\
\bottomrule
\end{tabular}%
}
\label{lorenz_table}
\end{table}

\subsection{Burgers' Equation}
\label{burger}
Burgers' equation is a non-linear PDE system that has been widely investigated due to its versatility in modeling various physical processes. It serves as a simplified framework for understanding phenomena such as fluid dynamics, gas dynamics, and traffic flow. Its mathematical structure, which balances nonlinear advection and viscous diffusion, makes it a widely used test case for developing and validating numerical methods.
The equation is given by:
\begin{equation}
    \frac{\partial u}{\partial t} +u \frac{\partial u}{\partial x} - \nu  \frac{\partial^2 u}{\partial x^2}=0,
\end{equation}
where $u(x,t)$ represents the velocity field, and $\nu$ denotes the viscosity coefficient. In this experiment, $\nu$ takes the value of 0.1. The training data for Burgers' equation is generated with 256 different positions spaced equally with $x\in[-8,8]$, and 101 different time steps spaced equally with $t\in [0,10]$. The candidate library consists of 10 terms with polynomial order up to 2, with the basis being $[1,u,u_x,u_{xx},u^2,uu_x,uu_{xx},u_x^2,u_xu_{xx},u_{xx}^2]$. The training data is injected with 0.2\% noise. For Burgers' equation, the initial condition is set to $u(0,x)=exp(-(x - 3)^2 / 2)$.

\begin{figure}[h]
    \centering
    \includegraphics[width=1\linewidth]{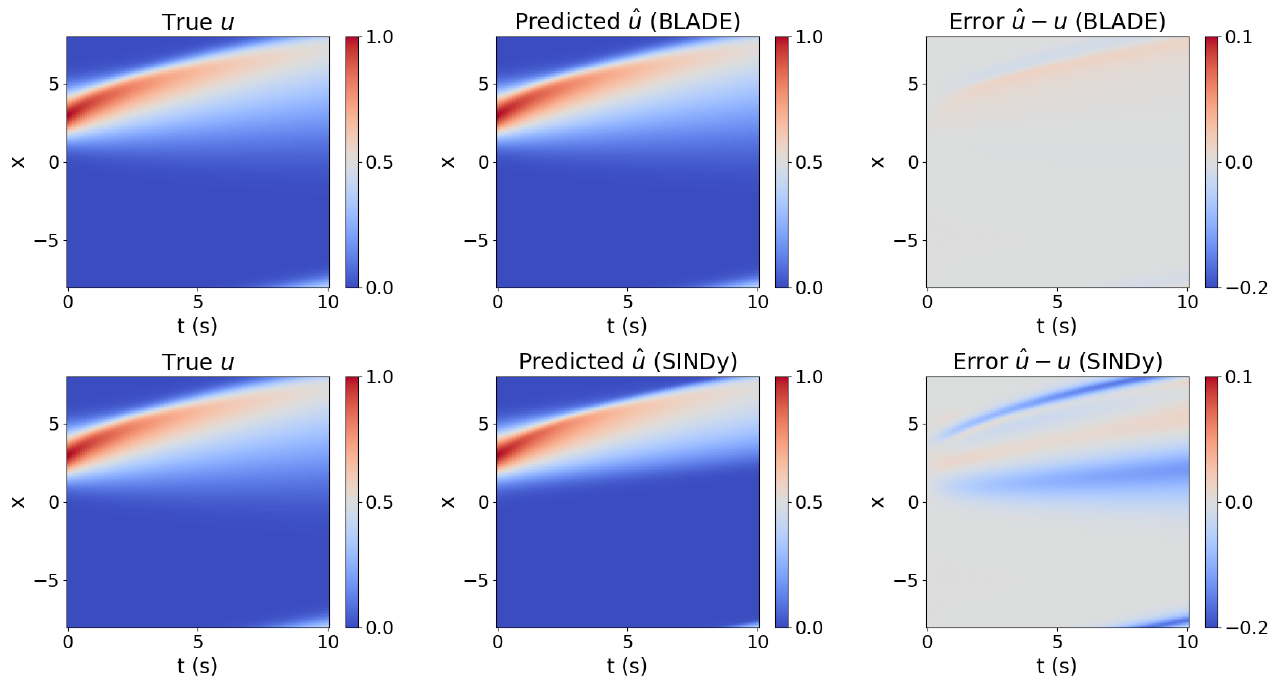}
    \caption{ Solution $u$ Contours of the identified Burgers' equation compared with ground truth (True $u$). \textbf{Top:} BLADE with default reSGLD as sampler \textbf{Bottom:} SINDy. \textbf{Left Column:} Contours of True $u$, \textbf{Middle Column:} Contours of predicted $\hat{u}$, \textbf{Right Column:} Contours of Error.}
    \label{fig:compare-burger}
\end{figure}

\begin{figure}[h]
    \centering
    \includegraphics[width=1\linewidth]{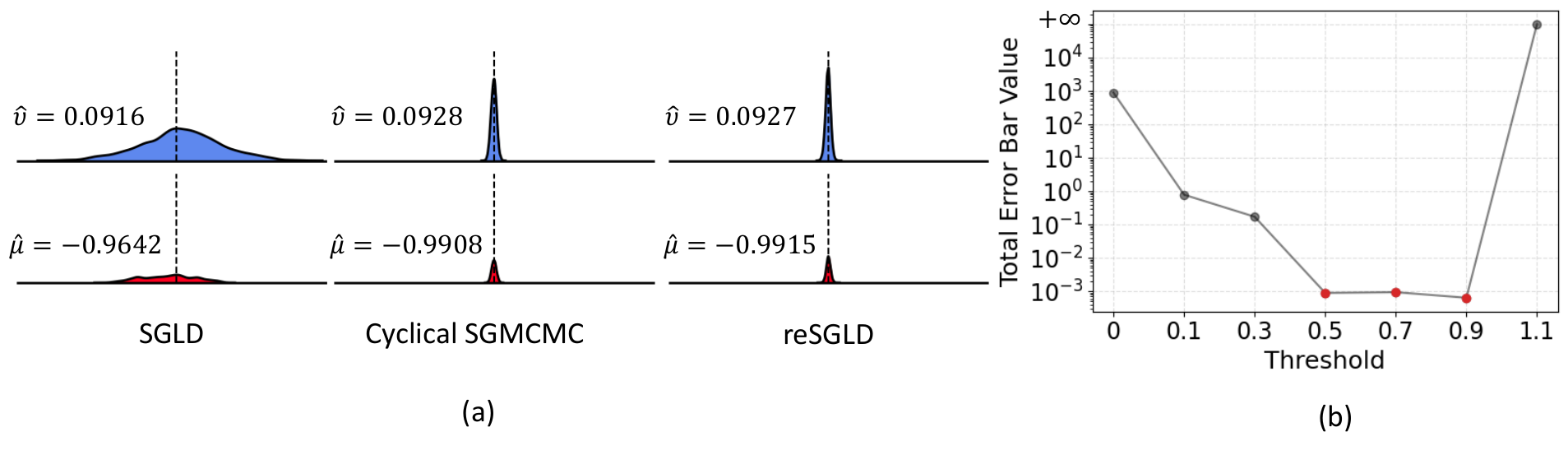}
    \caption{(a) Posterior distributions of identified model coefficients for Burgers' equation with different Langevin MCMC samplers adopted in BLADE. Here we denote the predicted coefficient for $u \frac{\partial u}{\partial x}$ as $\hat{\mu}$.  (b) An illustration of Total Error Bar (plotted in log scale) guiding threshold selection, $(0.5-0.7)$, with convection–diffusion equation.}
    \label{fig:burger-advection}
\end{figure}

Table \ref{table:burgers} presents the identified coefficients for the Burgers' equation using different methods. The results demonstrate better performance of BLADE compared to both SINDy and UQ-SINDy approaches. 
The SINDy method identified several spurious terms, including constant, linear ($u$), and nonlinear terms ($u^2$, $uu_{xx}$). Although UQ-SINDy shows improvement in MSE compared to SINDy, the identified system still contains a number of superfluous terms.
All three BLADE variants (SGLD, cyclical SGMCMC, and reSGLD) successfully identify the correct structure of the Burgers' equation, capturing both the nonlinear advection term ($uu_x$) and the diffusion term ($u_{xx}$) without introducing spurious terms. Among these, BLADE with reSGLD demonstrates the best performance with the lowest Total Error Bar (1.218$\cdot 10^{-4}$), MSE (1.435$\cdot 10^{-5}$), and AIC (-2.886$\cdot 10^{5}$).
Figure \ref{fig:compare-burger} provides a visual comparison between the true Burgers' equation and the systems identified by BLADE (reSGLD) and SINDy. 
The heatmaps illustrate the spatiotemporal evolution of the solution, where the prediction from BLADE demonstrates a closer correspondence to the ground truth than that of SINDy.
The error plot for BLADE reveals minimal discrepancies concentrated primarily in regions of larger magnitude, whereas SINDy exhibits larger and more widespread errors throughout the domain.
Figure \ref{fig:burger-advection}(a) displays the posterior distributions of the identified coefficients for the Burgers' equation using different Langevin MCMC samplers. The distributions show that reSGLD provides narrower, more concentrated posterior distributions compared to SGLD, indicating higher confidence in the estimated parameters. The cyclical SGMCMC method also yields well-concentrated posteriors, though slightly broader than those from reSGLD. This demonstrates that replica exchange sampling effectively enhances the exploration of the parameter space while maintaining precision in the identified coefficients.

\begin{table}[h]
\centering
\scriptsize
\caption{Learned coefficients with different models on Burgers' equation with 0.2\% Noise} 
\adjustbox{center=\textwidth}{%
\begin{tabular}{l cccccc}
\toprule
\multirow{2}{*}{\text{Basis}} 
& \multicolumn{6}{c}{\text{Burgers' equation}} \\
\cmidrule(lr){2-7}
 & \text{True} & \text{SINDy} & \text{UQ-SINDy}& \text{BLADE(SGLD)} & \text{BLADE(cyclical SGMCMC)} & \text{BLADE(reSGLD)} \\
\midrule
$1$ & - & - &1.315e-3& - & - & - \\
$u$ & - & 8.660e-2 &5.908e-3& - & - & - \\
$u_x$ & - & -2.195e-1&-1.363e-2 & - & - & - \\
$u_{xx}$ & 1.000e-1 & 5.431e-2&5.431e-2 & 9.161e-2 & \textbf{9.276e-2} & 9.274e-2 \\
$u^2$ & - & -1.960e-1&-3.109e-2 & - & - & - \\
$u u_x$ & -1.000e0 & -2.476e-1 &-9.494e-1& -9.642e-1 & -9.908e-1 & \textbf{-9.915e-1} \\
$u u_{xx}$ & - & 1.921e-1 &3.481e-2& - & - & - \\
$u_x^2$ & - & -&1.596e-2 & - & - & - \\
$u_x u_{xx}$ & - & - &-4.601e-2& - & - & - \\
$u_{xx}^2$ & - & - &7.756e-3& - & - & - \\
\midrule
\text{Total Error Bar} 
& - & - &3.503e-2& 2.983e-2 & 1.402e-4 & \textbf{1.218e-4} \\
\midrule
\text{MSE} 
& - & 1.933e-3 &3.490e-5& 3.619e-5 & 2.827e-5 & \textbf{1.435e-5} \\
\midrule
\text{AIC} 
& - & -1.15e5&-2.653e5 & -2.644e5 & -2.877e5 & \textbf{-2.886e5} \\
\bottomrule
\end{tabular}%
}
\label{table:burgers}
\end{table}

\subsection{Convection–Diffusion Equation}
\label{sec:4.4}
The convection–diffusion equation is a widely studied partial differential equation that describes the transport of a scalar quantity under the combined effects of convection and diffusion. It plays a critical role in modeling processes such as heat transfer, pollutant dispersion in fluids, and mass transfer in chemical engineering. The balance between convective transport, which moves the quantity with a flow, and diffusive transport, which acts to spread the quantity, makes it a fundamental equation for analyzing transport phenomena. The equation is expressed as:
\begin{equation}
    \frac{\partial u}{\partial t} +c \frac{\partial u}{\partial x} - D  \frac{\partial^2 u}{\partial x^2}=0,
\end{equation}
where the term $u(x,t)$ is the transported quantity, and the parameters $c$ and $D$ correspond to the convection velocity and the diffusion coefficient.
The training data for convection–diffusion equation is generated with 201 different positions spaced equally with $x\in[0,20]$, and 501 different time steps spaced equally with $t\in [0,5]$. The candidate library consists of 10 terms with polynomial order up to 2 with the basis being $[1,u,u_x,u_{xx},u^2,uu_x,uu_{xx},u_x^2,u_xu_{xx},u_{xx}^2]$. The training data is injected with 0.1\% noise.

\begin{figure}[h]
    \centering
    \includegraphics[width=1\linewidth]{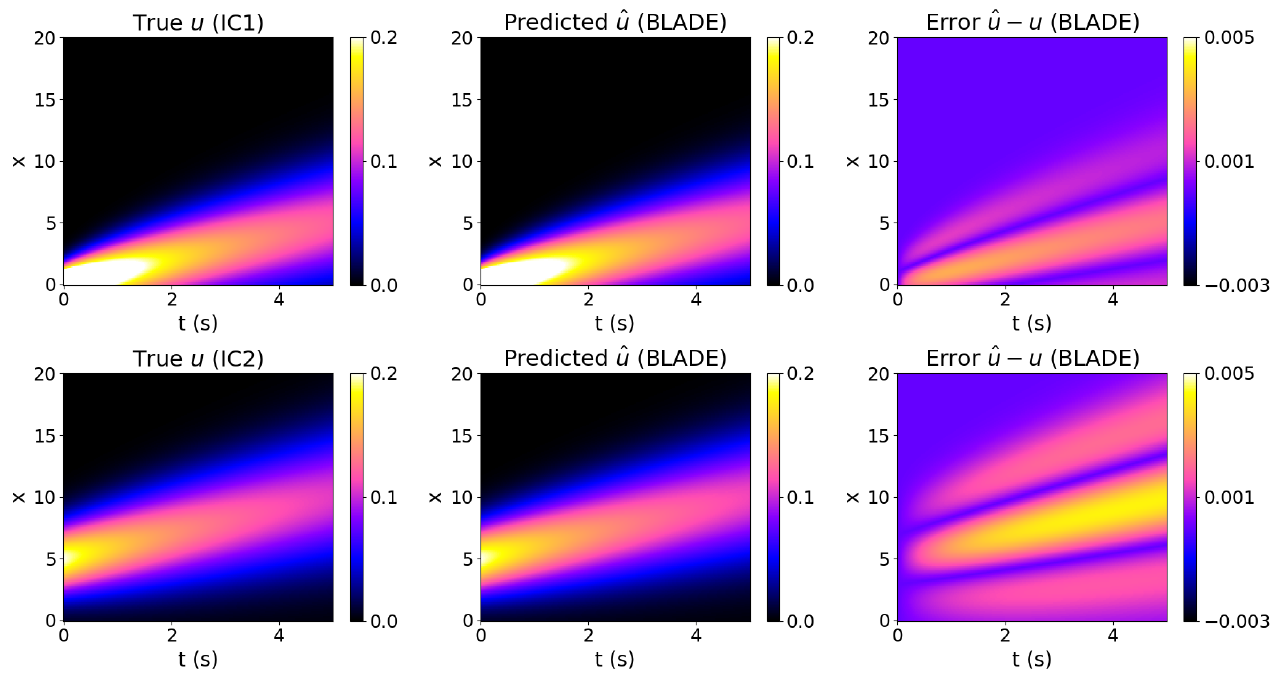}
    \caption{Solution $u$ Contours of the identified convection–diffusion equation using BLADE compared with ground truth (True $u$). \textbf{Top:} Initial condition IC1, \textbf{Bottom:} Initial condition IC2. \textbf{Left Column:} Contours of True $u$, \textbf{Middle Column:} Contours of predicted $\hat{u}$, \textbf{Right Column:} Contours of Error.}
    \label{fig:advection}
\end{figure}

In this experiment, we first discover the configuration of the thresholding settings. Figure \ref{fig:burger-advection}(b) illustrates how the Total Error Bar value changes with different threshold settings for the convection–diffusion equation. The clear U-shaped curve indicates an optimal balance between model complexity and accuracy. At lower thresholds, the model includes too many terms, while at higher thresholds, essential terms are excluded, driving the model to learn poorly with a higher standard deviation for coefficients. The Total Error Bar reaches its minimum in the range of 0.5 to 0.9, providing a data-driven approach to threshold selection in the sequential thresholding process. Subsequently, Table \ref{table:advection} compares the performance of SINDy, UQ-SINDy, and the proposed BLADE method in identifying the convection–diffusion equation under 0.1\% noise. The BLADE framework achieves the closest alignment with the true coefficients (e.g., $u_x: -1.000$, $u_{xx}: 0.960$), while other methods exhibit significant deviations or spurious terms (e.g., erroneous $u u_x$ or $u_{xx}^2$ coefficients). BLADE also demonstrates superior robustness, with the lowest MSE ($1.929 \times 10^{-5}$), minimal total error bar ($8.613 \times 10^{-4}$), and optimal AIC ($-1.093 \times 10^6$), underscoring its efficacy in sparse, noise-resilient dynamics discovery. Figure~\ref{fig:advection} compares the ground-truth and predicted solutions under different initial conditions, along with the corresponding error maps, showing that the predictions closely match the ground truth.

\begin{table}[h]
\centering
\scriptsize
\caption{Learned coefficients with different models on convection–diffusion equation with 0.1\% noise}
\adjustbox{center=\textwidth}{
\begin{tabular}{l cccccc}
\toprule
\multirow{2}{*}{\text{Basis}} 
& \multicolumn{6}{c}{\text{convection–diffusion equation}} \\
\cmidrule(lr){2-7}
 & \text{True} & \text{SINDy} &\text{UQ-SINDy}& \text{BLADE (SGLD)} & \text{BLADE (cyclical SGMCMC)} & \text{BLADE (reSGLD)} \\
\midrule
$1$ & - & -&4.090e-4 & - & - & - \\
$u$ & - & -&4.051e-2 & - & - & - \\
$u_x$ & -1.000e0 & -1.265e0&-1.200e0 & -9.985e-1 & -1.012e0  & \textbf{-1.000e0} \\
$u_{xx}$ & 1.000e0 & -&4.686e-1 & 9.514e-1 & 9.551e-1 & \textbf{9.600e-1} \\
$u^2$ & - & - &-6.343e-1& - & - & - \\
$u u_x$ & - & 1.291e0&1.283e-0 & - & - & - \\
$u u_{xx}$ & - & 5.544e-1 &1.460e0& - & - & - \\
$u_x^2$ & - & -&1.148e0 & - & - & - \\
$u_x u_{xx}$ & - & -&-8.611e-3 & - & - & - \\
$u_{xx}^2$ & - & 3.627e-1&9.536e-1 & - & - & - \\
\midrule
\text{Total Error Bar} 
& - & -&1.123e1 & 5.141e-3 & 1.486e-3 & \textbf{8.613e-4} \\
\midrule
\text{MSE} 
& - & 3.522e-4&2.053e-5 & 1.944e-5 & 1.942e-5 & \textbf{1.929e-5} \\
\midrule
\text{AIC} 
& - & -8.006e5&-1.087e6 & -1.089e6 & - 1.092e6 & \textbf{-1.093e6} \\
\bottomrule
\end{tabular}}
\label{table:advection}
\end{table}

\section{Data-Efficient System Discovery with Active Learning}
Having demonstrated the robustness of BLADE with sufficient data in Section \ref{section:experiment}, we now evaluate its performance in data-scarce scenarios with AL. Given the limited complexity of the sample size inherent to AL settings, we employ the \textit{Metropolis adjusted Langevin algorithm} (MALA) with exact gradients via BLADE, which serves as a computationally efficient single-chain alternative to reSGLD \citep{chewi2021optimal,dwivedi2019log}.

\subsection{Lotka-Volterra System with AL}

We evaluate the BLADE framework with AL on the Lotka-Volterra system, following the procedure outlined in Algorithm \ref{al:activelearning}. The pool of potential design points for this experiment, denoted $\mathcal{D}$, consists of $N=10,000$ points randomly sampled from the states generated in the 50,000 time steps. The data simulation procedure is the same as in Section \ref{lotka_volterra}, except that $50,000$ time steps of states are collected, with $\Delta t=5\cdot 10^{-4}$. Furthermore, 5\% additive noise is injected during experiments. With $\mathcal{D}$ being randomly sampled from the simulated data, it can be viewed as an irregularly sampled data set. 
In this synthetic experiment setting, the acquisition of the time derivative is carried out using the original regularly sampled data. However, in realistic conditions, the acquisition of the response variable (the time derivative) is computationally expensive due to the irregularity of the available data. Here $\lambda=0,\alpha=0.5$ are adopted.

\begin{figure}[h]
    \centering
    \includegraphics[width=1\linewidth]{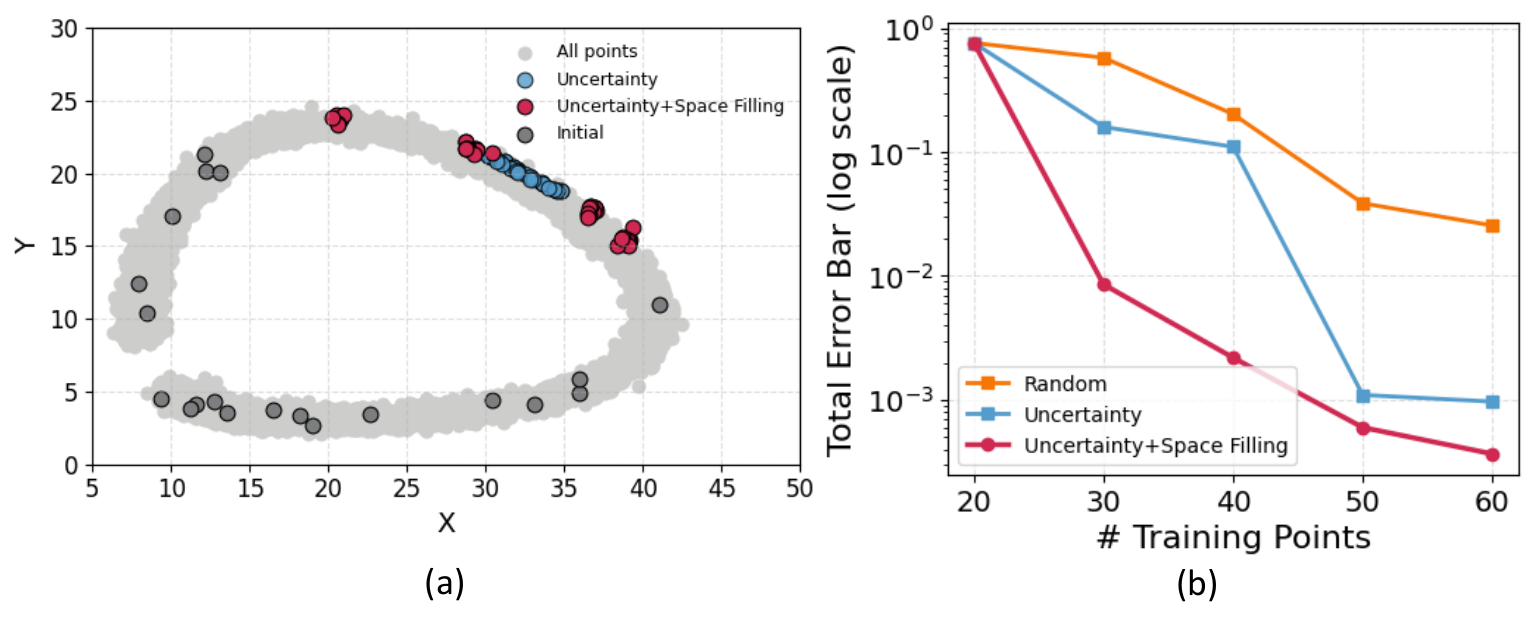}
    \caption{Illustration of active learning for the BLADE method on the Lotka–Volterra system. (a) All potential sample points (light gray) and the initial 20 randomly chosen training points (dark gray) for the first round. Red points show points added over the four following iterations using the Uncertainty + Space-Filling acquisition function, whereas blue points reflect the purely Uncertainty-based selection.
(b) Log-scale plot of the Total Error Bar for Random, Uncertainty, and Uncertainty + Space-Filling acquisition functions as training points are incrementally added.}
    \label{fig:al}
\end{figure}

Training begins with an initial set of 20 randomly selected points. In each subsequent round, 10 additional points are selected via the specified acquisition functions and added to $\mathcal{M}$. Based on the acquisition function of Uncertainty+Space Filling, the tolerance of convergence $\textit{Tol}$ is set to 0.8, and experiments with other acquisition functions are carried out for the same number of rounds. As depicted in Figure \ref{fig:al}(a), combining space filling and uncertainty criteria yields a more uniform spatial distribution of the sampled points, while relying solely on uncertainty concentrates the sampling in regions of high state magnitude. Clustered data in one area have a higher chance of model overfitting and lack representativeness of the whole dataset. As shown in Figure \ref{fig:al}(b) and Table \ref{lotka-al}, although the Total Error Bar in the round with 60 points does not differ much ($\Delta = 6\cdot 10^{-3} $ on linear scale), the learned system combining Uncertainty and Space Filling yields a learned system more closely aligned with the true dynamics. Furthermore, this combined strategy illustrates the efficiency of systematic AL over random sampling. For the Total Error Bar to drop below $10^{-2}$, the random sampling strategy requires 70 data, while the combined strategy uses 30 data, reducing required measurements by approximately 60\%. AL acquisition functions enhance the informativeness and representativeness of selected points, other than training efficiency, under the same round of training; it also improves alignment with the ground truth dynamics. As shown in Table \ref{lotka-al}, compared to randomly adding points, the Uncertainty and Uncertainty+Space-Filling strategies more accurately preserve the nonlinear structure of the ground truth, capturing both growth and predator–prey interactions without extra terms.

\begin{table}[h!]
    \centering
    \scriptsize
    \caption{Comparison of true system (Lotka-Volterra) and identified systems for different strategies with 60 training points. Our method (Uncertainty+Space-Filling) is shown \textbf{bolded}.}
    \adjustbox{center=\textwidth}{
    \begin{tabular}{@{}lcc@{}}
        \toprule
        \textbf{Methods} & \boldmath{$x_t$} & \boldmath{$y_t$} \\
        \midrule
        True system & $x_t = x - 0.1xy$ & $y_t = -1.5y + 0.075xy$ \\
        Random & $x_t = 0.181 + 0.992x - 0.099xy$ & $y_t = -0.311 - 1.449y + 0.073xy$ \\
        Uncertainty & $x_t = 0.902x - 0.091xy$ & $y_t = -1.339y + 0.067xy$ \\
        \textbf{Uncertainty+Space-Filling} & \boldmath{$x_t = 0.998x - 0.096xy$} &\boldmath{$y_t = -1.423y + 0.070xy$} \\
        \bottomrule
        \end{tabular}}
    \label{lotka-al}
\end{table}

\subsection{Burgers' Equation with AL}
\label{sec_burger_al}

We next evaluate BLADE with AL on Burgers' equation. The original data here distributed approximately 4,000 positions equally spaced in [-8,8] and 1,000 different time steps equally spaced in [0,10]. The initial condition is set to the same as in Section \ref{burger}. 
The pool of potential design points is constructed by selecting three distinct time steps (1s, 5s, and 8s) from the original data. These time points serve as a representative random sample, and any arbitrary selection of time steps from the dataset would be equally suitable.
Under this scenario, the calculation of partial derivatives with respect to $t$ can be difficult to obtain, hence in need of active learning. Again, in this synthetic data scenario, we calculate $\mathbf{U}_t$ based on the original data. The candidate library adopted in this experiment is the same as in Section \ref{burger} with 10 terms $[1,u,u_x,u_{xx},u^2,uu_x,uu_{xx},u_x^2,u_xu_{xx},u_{xx}^2]$. Here we inject i.i.d. noise on $\dot{\mathbf{U}}$ following $lognormal\,(0,0.1)$.

\begin{figure}
    \centering
    \includegraphics[width=0.5\linewidth]{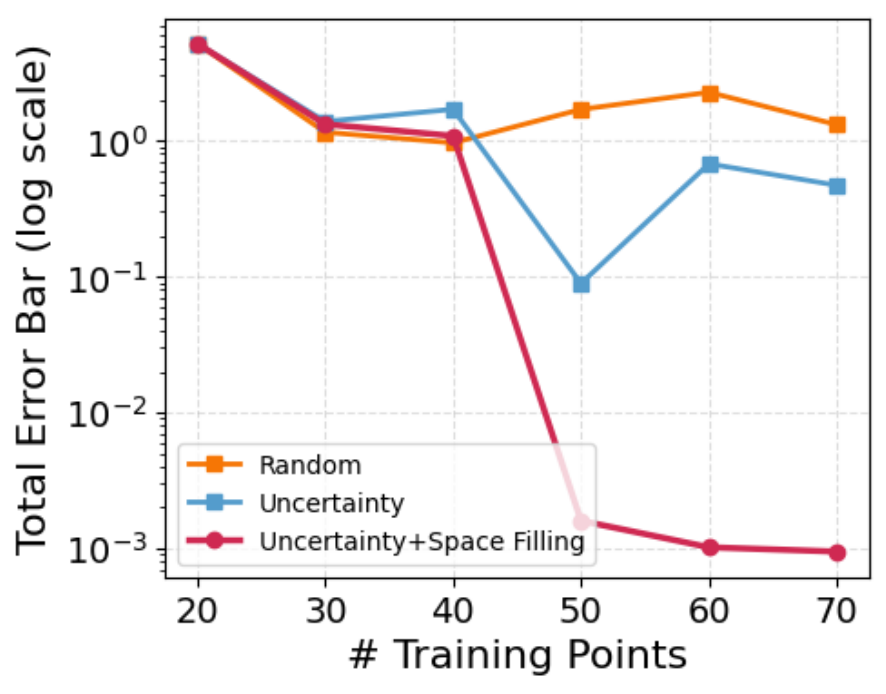}
    \caption{Active Learning on Burgers' equation: Log-scale plot of the Total Error Bar for Random, Uncertainty, and Uncertainty + Space-Filling acquisition functions as training points are incrementally added.}
    \label{fig:burgers-al}
\end{figure}

Training begins with an initial set of 20 randomly selected data. In each subsequent round, 10 additional data are selected via the specified acquisition functions and added to $\mathcal{M}$. The tolerance of convergence $\textit{Tol}$ is set to 0.3, and experiments with other acquisition functions are carried out for the same number of rounds as the uncertainty+space-filling acquisition function. 
Following the initial training phase, the values of $d(\mathbf{u})$ ranged from a minimum of $6.769 \times 10^{-11}$ to a maximum of $1.443$. This substantial relative difference, exceeding ten orders of magnitude, prompted a density adjustment with parameters set to $\lambda=0.5, \alpha =0.3$.
As shown in Figure~\ref{fig:burgers-al}, AL approach utilizing a hybrid acquisition function demonstrates enhanced data efficiency, achieving strong performance with a reduced number of training points.
The hybrid method identifies the correct candidate functions with 50 points, while the uncertainty method and the random method are left with 4 and 7 basis functions, respectively.
For the Total Error Bar to drop below $10^{-2}$, the random sampling strategy requires 90 data, whereas the hybrid strategy only requires 50 data, which reduces the required measurements by 44\%. As shown in Table \ref{table: burgers-al}, the random sampling approach results in an equation contaminated with extraneous terms such as $u, u^2, uu_x, u_x u_{xx}$, which deviate significantly from the true system. This suggests that randomly selected training points may not be able to adequately capture the underlying physics, leading to overfitting or the inclusion of spurious terms. On the other hand, uncertainty-based sampling shows improvement by reducing the number of incorrect terms. However, the coefficient estimates remain suboptimal, particularly for the non-linear interaction term $u \frac{\partial u}{\partial x}$. This indicates that an uncertainty-only acquisition function may still suffer from biased selection, potentially over-exploring regions with high variance while neglecting global representativeness. The hybrid uncertainty+space-filling strategy demonstrates the most accurate identification of the governing equations. The identified terms match the true system structure, and the coefficient values closely approximate their theoretical counterparts. This highlights the advantage of balancing exploration (uncertainty) and diversity (space-filling) when selecting training points, leading to more robust equation discovery with fewer data points.

\begin{table}[h!]
    \centering
    \scriptsize
    \caption{Comparison of true system (Burgers' equation) and identified systems for different strategies with 70 training data. Our method (Uncertainty+Space-Filling) is shown \textbf{bolded}.}
    \adjustbox{center=\textwidth}{
    \begin{tabular}{@{}lcc@{}}
        \toprule
        \textbf{Methods} & \textbf{Learned systems} \\
        \midrule
        True system & $\frac{\partial u}{\partial t} +u \frac{\partial u}{\partial x} - 0.1  \frac{\partial^2 u}{\partial x^2}=0$  \\
        Random & $\frac{\partial u}{\partial t}-0.054u-0.075\frac{\partial u}{\partial x}+0.147\frac{\partial^2 u}{\partial x^2}+0.944u^2+0.060u\frac{\partial u}{\partial x}+0.115\frac{\partial u}{\partial x}\frac{\partial^2 u}{\partial x^2}=0$ \\
        Uncertainty & $\frac{\partial u}{\partial t}+0.14\frac{\partial u}{\partial x}+0.714u\frac{\partial u}{\partial x}-0.073\frac{\partial^2 u}{\partial x^2}=0$  \\
        \textbf{Uncertainty+Space-Filling} & \boldmath{$\frac{\partial u}{\partial t}+0.971u\frac{\partial u}{\partial x}-0.090\frac{\partial^2 u}{\partial x^2}=0$} \\
        \bottomrule
        \end{tabular}}
    \label{table: burgers-al}
\end{table}

\subsection{Convection-Diffusion Equation with AL}
We further evaluate BLADE with active learning on the convection--diffusion equation. We randomly select 600 points from the full dataset in Section~\ref{sec:4.4} to form the pool of potential design points, denoted by \(\mathcal{D}\). Because these points are sampled irregularly in \((x,t)\), accurate estimation of partial derivatives with respect to \(x\) and \(t\) becomes more challenging, which highlights the need for active learning. As in the previous synthetic-data experiments, we compute \(\mathbf{U}_t\) using the original data from Section~\ref{sec:4.4}. The candidate library is the same as in Section~\ref{sec:4.4}, consisting of 10 terms:
$[1,u,u_x,u_{xx},u^2,uu_x,uu_{xx},u_x^2,u_xu_{xx},u_{xx}^2]$. We additionally perturb \(\mathbf{U}\) with i.i.d. additive Gaussian noise at a level of \(0.1\%\).

\begin{figure}
    \centering
    \includegraphics[width=1\linewidth]{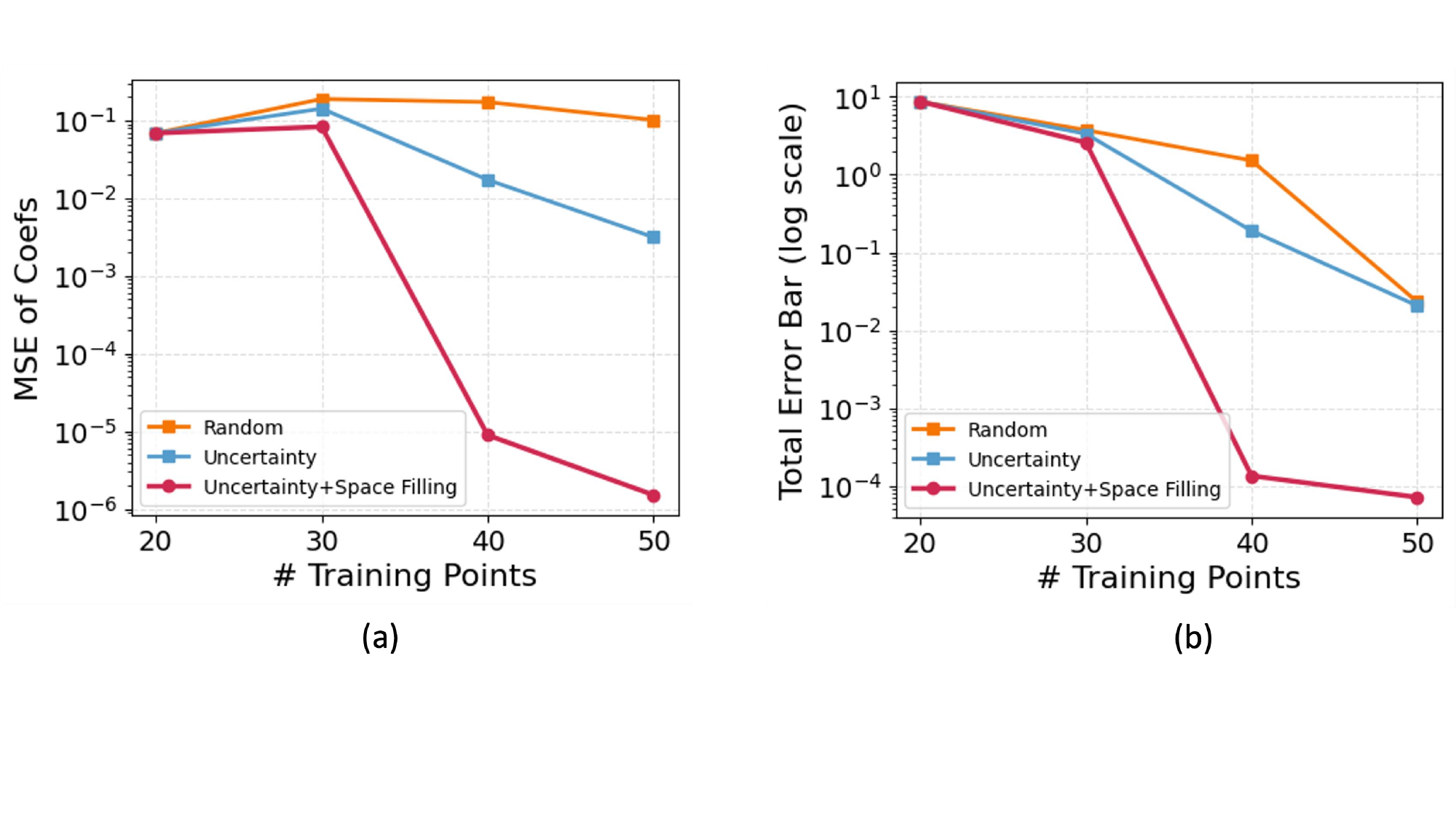}
    \caption{Illustration of active learning for the BLADE method on the Convection-Diffusion equation. (a) Log-scale plot of the MSE of coefficients for Random, Uncertainty, and Uncertainty + Space-Filling acquisition functions as training points are incrementally added.
(b) Log-scale plot of the Total Error Bar for Random, Uncertainty, and Uncertainty + Space-Filling acquisition functions as training points are incrementally added.}
    \label{fig:advection-al}
\end{figure}

Training starts from an initial set of 20 randomly selected data points. At each subsequent active learning round, 10 additional points are selected by the specified acquisition function and added to \(\mathcal{M}\). The tolerance of convergence is fixed at \(\mathit{Tol}=0.3\). To ensure a fair comparison, all acquisition strategies are run for the same number of rounds as the uncertainty+space-filling strategy. For the hybrid acquisition function, we use the density-adjustment parameters \(\lambda=0.5\) and \(\alpha=0.3\).
Figure~\ref{fig:advection-al} shows that the hybrid acquisition strategy (uncertainty+space-filling) is substantially more data-efficient than either random sampling or uncertainty-only sampling. In Fig.~\ref{fig:advection-al}(a), we report the coefficient error, measured as the mean squared error (MSE) between the identified PDE coefficients and the ground-truth coefficients. A slight increase in coefficient MSE is observed from 20 to 30 training points for all three strategies. This behavior is expected in sequential active discovery: early queried points are chosen to improve informativeness and coverage, which can temporarily shift the posterior and the selected sparse support before the coefficient estimates stabilize. As more points are added, the coefficient MSE decreases for all methods, but the reduction is much sharper for the hybrid strategy, dropping by 5 orders of magnitude and reaching the lowest error among all methods.
Figure~\ref{fig:advection-al}(b) reports the total error bar. The hybrid strategy shows the fastest reduction and achieves a near-converged value by 40 training points, while the random and uncertainty-only strategies remain substantially larger at the same budget. This behavior is consistent with the identified systems in Table~\ref{table: addvection-al}: with 50 training points, the hybrid strategy recovers the correct convection--diffusion structure with coefficients very close to the ground truth (\(1.002\) for \(u_x\) and \(-0.996\) for \(u_{xx}\)), whereas the random strategy misses the diffusion term \(u_{xx}\) entirely, and the uncertainty-only strategy identifies the correct basis but with noticeably larger coefficient error. Overall, these results highlight the benefit of combining exploration (uncertainty) with diversity (space-filling): the hybrid acquisition function improves support recovery and coefficient accuracy using fewer training points, leading to more robust equation discovery in the irregularly sampled setting.

\begin{table}[h!]
    \centering
    \scriptsize
    \caption{Comparison of true system (Convection-Diffusion equation) and identified systems for different strategies with 50 training data. Our method (Uncertainty+Space-Filling) is shown \textbf{bolded}.}
    \adjustbox{center=\textwidth}{
    \begin{tabular}{@{}lcc@{}}
        \toprule
        \textbf{Methods} & \textbf{Learned systems} \\
        \midrule
        True system & $\frac{\partial u}{\partial t} +\frac{\partial u}{\partial x} - \frac{\partial^2 u}{\partial x^2}=0$  \\
        Random & $\frac{\partial u}{\partial t} +0.869\frac{\partial u}{\partial x}=0$ \\
        Uncertainty & 
        $\frac{\partial u}{\partial t}+0.767\frac{\partial u}{\partial x}-0.662\frac{\partial^2 u}{\partial x^2}=0$
        \\
        \textbf{Uncertainty+Space-Filling} & \boldmath{$\frac{\partial u}{\partial t}+1.002\frac{\partial u}{\partial x}-0.996\frac{\partial^2 u}{\partial x^2}=0$} \\
        \bottomrule
        \end{tabular}}
    \label{table: addvection-al}
\end{table}

\section{Conclusion and Discussion}
This work presents a novel Lagevin-assisted Bayesian active learning framework, BLADE, which integrates reSGLD with AL to discover governing equations from noisy and limited experimental data. Our comprehensive evaluation across diverse dynamical systems demonstrates that incorporating reSGLD sampling within the physical discovery process provides significant advantages over traditional approaches, especially in the face of noise. By simultaneously operating multiple Markov chains at different temperatures, BLADE achieves a better exploration of the parameter space, effectively navigating the complex landscapes of nonlinear dynamical systems. Traditional sparse regression methods for physical discovery (like SINDy) provide point estimates of system coefficients but offer limited insight into model UQ. In contrast, BLADE offers robust UQ for both model parameters and system trajectories. Moreover, while sparsity-inducing priors (as in UQ-SINDy) can theoretically promote sparse solutions, our experiments improved robustness and accuracy when identifying system dynamics and estimating the uncertainties. Furthermore, our AL extension introduces a hybrid acquisition function that balances uncertainty reduction with domain coverage. This approach significantly outperforms both random sampling and pure uncertainty-based acquisition, reducing measurement requirements by 44-57\% across test cases.

While BLADE demonstrates strong performance across benchmarks, several challenges remain. First, the computational cost of reSGLD sampling exceeds that of deterministic optimization approaches. For very large systems or real-time applications, this overhead may become prohibitive. Future work would explore algorithmic optimizations, like adaptive temperature scheduling or parallel chain implementation.
Second, our current implementation determines the initial library of candidate functions a priori. For complex systems where this library is incomplete or unknown, incorporating neural operators or manifold-learned feature spaces may enable automatic basis discovery.
Third, BLADE presently relies on finite-difference derivative estimates, which can be fragile in noisy PDE settings.
Pointwise numerical differentiation is well known to amplify measurement noise, which can degrade the candidate library construction and the downstream system-identification step.
Improving robustness in high-noise PDE settings is an important direction for future work. One promising approach is to extend BLADE to weak-form or integral formulations, which avoid explicit pointwise derivative estimation by transferring derivatives from noisy data to smooth test functions and thus naturally reduce noise amplification \citep{messenger2021weak, messenger2020weak}. Another effective strategy is transforming the candidate library from the time/spatial domain into the Laplace or frequency domain. Methods such as LES-SINDy \citep{zheng2024les_sindy} employ the Laplace transform alongside integration by parts to evaluate derivatives analytically in the transformed space, effectively handling noisy measurements and high-order derivatives without direct numerical differentiation. Similarly, filtering techniques based on the Discrete Fourier Transform (DFT) can be applied to isolate and attenuate noise-dominated frequencies prior to candidate library construction \citep{thanasutives2022noiseaware}. A third direction is to incorporate neural-network-based denoising or surrogate representations and compute derivatives via automatic differentiation, potentially combined with physics-informed or time-stepping-constrained architectures to better separate measurement noise from the underlying dynamics \citep{rudy2019deep, thanasutives2022noiseaware}.
These extensions would decouple BLADE's Bayesian inference and uncertainty-quantification components from the current finite-difference preprocessing bottleneck, and could substantially improve robustness in strongly noisy regimes.

In summary, BLADE constitutes a robust, uncertainty-aware, and data-efficient framework for the discovery of interpretable dynamical systems. By bridging Bayesian inference, stochastic sampling, and AL, it addresses fundamental challenges in data-driven science. As high-fidelity experimental data remain costly and limited, the principles of Bayesian active discovery embodied in BLADE are expected to play an increasingly central role in the next generation of data-driven physics and engineering.

\section*{Declaration of Competing Interest}
The authors declare that they have no known competing financial interests or personal relationships that could have appeared to
influence the work reported in this paper.

\section*{Data Availability}
Data will be made available on request from the authors.

\section*{Funding}
We would like to thank the support of National Science Foundation (DMS-2533878, DMS-2053746, DMS-2134209, ECCS-2328241, CBET-2347401 and OAC-2311848), and U.S.~Department of Energy (DOE) Office of Science Advanced Scientific Computing Research program DE-SC0023161, the SciDAC LEADS Institute, and DOE–Fusion Energy Science, under grant number: DE-SC0024583.

\appendix

\section{Replica-exchange hyperparameter specification and swap-rate behavior}
\label{app:swap_rate}

\begin{figure}[t]
    \centering
    \includegraphics[width=0.72\textwidth]{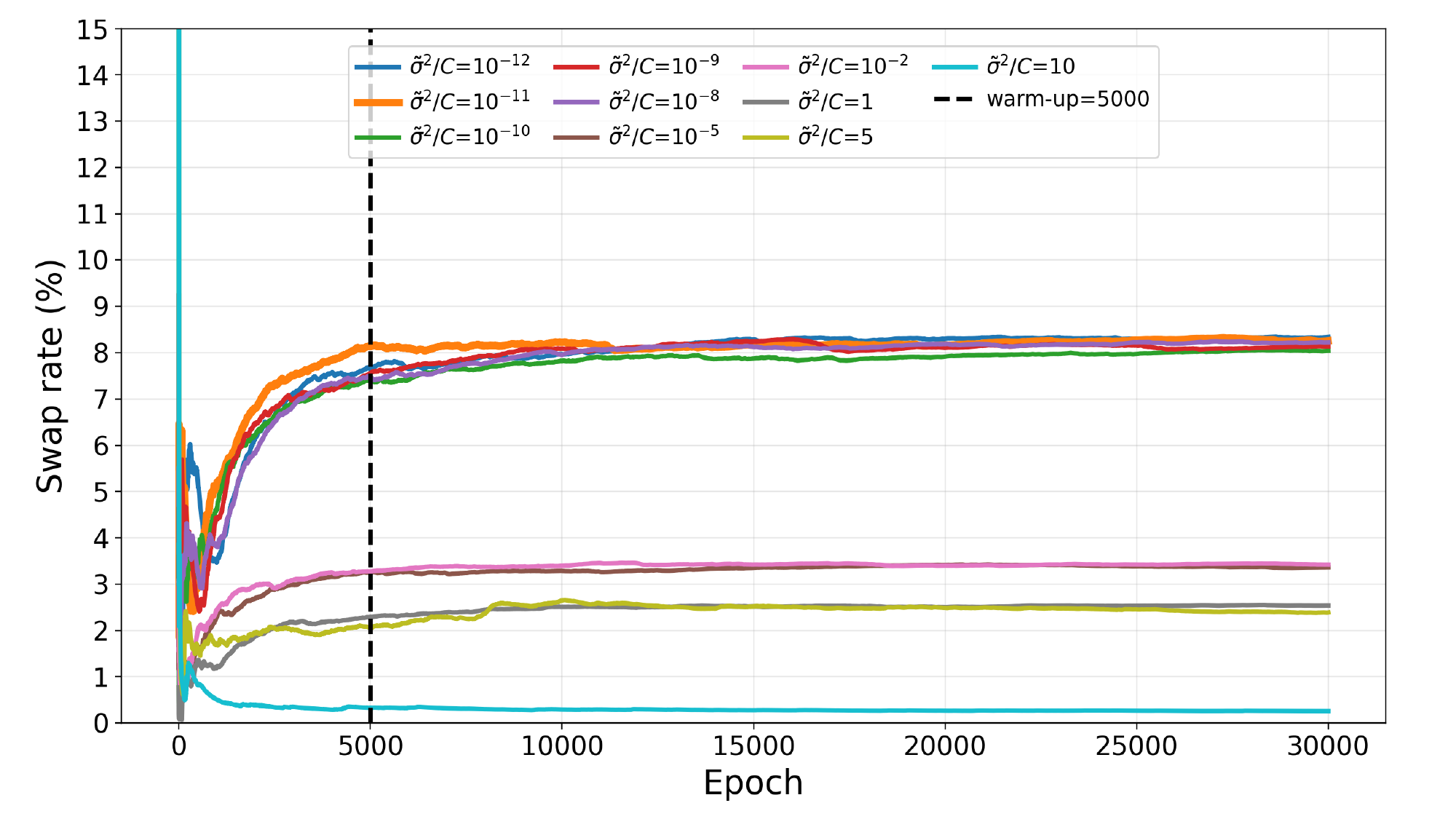}
    \caption{Swap acceptance rate over algorithmic iterations for representative values of \(\tilde{\sigma}^2/C\) on the Burgers' benchmark (first round). The dashed vertical line indicates the warm-up boundary at 5000 iterations. Values of \(\tilde{\sigma}^2/C\) around $10^{-11}$ produce stable post--warm-up swap rates in a practically usable range, whereas overly large values substantially suppress exchange.}
    \label{fig:swap-rate-hatvar}
\end{figure}

In the reSGLD sampler used in BLADE, we fix the two temperatures in Eq.~\eqref{eq:resgld iteration} as $\tau_1 = 1, \tau_2 = 2$
and treat the ratio \(\tilde{\sigma}^2/C\) in the corrected swapping function Eq.~\eqref{eq:swap rate} as an effective hyperparameter. Following prior reSGLD implementations \citep{deng2020non, zheng2024constrained}, we tune the ratio \(\tilde{\sigma}^2/C\) directly, rather than estimating \(\tilde{\sigma}\) and choosing \(C\) separately.
In the experiments, we first conduct a swap-rate ablation on the Burgers equation, where we plot the swap acceptance rate over algorithmic iterations for representative values of \(\tilde{\sigma}^2/C\); the dashed vertical line in Fig.~\ref{fig:swap-rate-hatvar} marks the warm-up boundary at 5000 iterations. This experiment is used to illustrate how the choice of \(\tilde{\sigma}^2/C\) affects exchange efficiency in practice, and values around \(10^{-11}\) lead to stable post--warm-up swap rates in a practically usable range for Burgers. We follow the same tuning principle in the other benchmarks, with \(\tilde{\sigma}^2/C\) set to \(6\) for the Lotka--Volterra system, \(3\) for the Lorenz system, and \(10^{-11}\) for both the Burgers and convection--diffusion equations, while fixing the replica temperatures to \(\tau_1=1\) and \(\tau_2=2\) throughout.


\paragraph{Tuning procedure.}
The choice of \(\tilde{\sigma}^2/C\) is made empirically based on the observed swap acceptance behavior. Practically useful settings typically yield post--warm-up swap acceptance rates in a moderate range of approximately \(5\%\) to \(15\%\). In practice, we begin with a coarse scan over a wide range of candidate values, typically starting from \(10\) and decreasing by orders of magnitude, in order to identify a regime in which replica exchange is neither degenerate nor overly suppressed. We then make local adjustments within that regime when needed for a given benchmark. As a result, selecting \(\tilde{\sigma}^2/C\) involves a modest amount of empirical tuning in practice.

\paragraph{Effect of \(\tilde{\sigma}^2/C\) on replica-exchange dynamics.}
Figure~\ref{fig:swap-rate-hatvar} shows the swap acceptance rate over algorithmic iterations for representative values of \(\tilde{\sigma}^2/C\) on the Burgers benchmark during the first round of coefficient selection, with the dashed vertical line marking the warm-up boundary at 5000 iterations. The main observation is that replica-exchange behavior remains relatively stable for values near \(10^{-11}\), where the post--warm-up swap acceptance rate stays in a moderate range. This suggests that the practical tuning of \(\tilde{\sigma}^2/C\) is not overly sensitive within a local neighborhood of the selected value. By contrast, substantially larger values suppress exchange more strongly and lead to noticeably lower post--warm-up swap rates. For the Burgers' equation benchmark, values of \(\tilde{\sigma}^2/C\) above \(10^{-8}\) lead to relatively inaccurate estimation of the coefficient of \(u_{xx}\), causing this term to be omitted and thereby resulting in incorrect system discovery. Overall, this behavior supports our empirical choice of \(\tilde{\sigma}^2/C\) and indicates that the selected value lies in a stable operating range for replica exchange.

\bibliography{refs}

\end{document}